\documentclass{article} 
\usepackage{iclr_r2fm,times}

\usepackage{amsmath,amsfonts,bm}

\def\eqref#1{equation~\ref{#1}}
\def\1{\bm{1}}

\DeclareMathAlphabet{\mathsfit}{\encodingdefault}{\sfdefault}{m}{sl}
\SetMathAlphabet{\mathsfit}{bold}{\encodingdefault}{\sfdefault}{bx}{n}

\usepackage[export]{adjustbox}
\usepackage[utf8]{inputenc} 
\usepackage[T1]{fontenc}    
\usepackage{url}            
\usepackage{booktabs}    
\usepackage{amsfonts}     
\usepackage{nicefrac}       
\usepackage{microtype}      
\usepackage{xcolor}
\usepackage[pagebackref,breaklinks,colorlinks,citecolor=blue]{hyperref}
\usepackage[ruled,vlined,linesnumbered]{algorithm2e}
\usepackage{amsmath}
\usepackage{multirow}
\usepackage{multicol}

\usepackage{amssymb}
\usepackage{longtable}
\usepackage{graphicx}
\usepackage{csvsimple}
\usepackage{rotating}
\usepackage{multirow}
\usepackage{diagbox}
\usepackage[most]{tcolorbox}

\iclrfinalcopy
\author{
Chen Cheng${}^{1*}$ \qquad
Xinzhi Yu${}^{2*}$ \qquad
Haodong Wen${}^{3}$\thanks{Equal contribution.} \qquad
Jingsong Sun${}^{3}$ 
\vspace{5pt}
\\
\vspace{5pt}
~\textbf{Guanzhang Yue${}^{4}$ \qquad
Yihao Zhang${}^{4}$\qquad
Zeming Wei${}^{4}$}\thanks{Corresponding author: Zeming Wei (weizeming@stu.pku.edu.cn).}\\
${}^{1}$ShanghaiTech University~
${}^{2}$Fudan University~
${}^{3}$Xi'an Jiaotong University~
${}^{4}$Peking University
}

\title{
Exploring the Robustness of\\ In-Context Learning with Noisy Labels
}

\begin{document}

\maketitle

\begin{abstract}
Recently, the mysterious In-Context Learning (ICL) ability exhibited by Transformer architectures, especially in large language models (LLMs), has sparked significant research interest. However, the resilience of Transformers' in-context learning capabilities in the presence of noisy samples, prevalent in both training corpora and prompt demonstrations, remains underexplored. In this paper, inspired by prior research that studies ICL ability using simple function classes, we take a closer look at this problem by investigating the robustness of Transformers against noisy labels. Specifically, we first conduct a thorough evaluation and analysis of the robustness of Transformers against noisy labels during in-context learning and show that they exhibit notable resilience against diverse types of noise in demonstration labels. Furthermore, we delve deeper into this problem by exploring whether introducing noise into the training set, akin to a form of data augmentation, enhances such robustness during inference, and find that such noise can indeed improve the robustness of ICL. Overall, our fruitful analysis and findings provide a comprehensive understanding of the resilience of Transformer models against label noises during ICL and provide valuable insights into the research on Transformers in natural language processing. Our code is available at \url{https://github.com/InezYu0928/in-context-learning}.
\end{abstract}

\section{Introduction}

In recent years, Large Language Models (LLMs) have achieved significant success across various tasks in real-world applications. Transformer~\citep{vaswani2023attention}, as the typical backbone architecture, also emerges an intriguing ability known as In-Context Learning (ICL)~\citep{brown2020language,dong2023survey} that the model can learn a new task with only a few input-output pairs {demonstrated} during inference \emph{without} modifying any model parameters. This mysterious property of transformers has attracted great research interests in understanding this ability, among which one representative perspective is investigating ICL through simple function classes~\citep{garg2023transformers}.

{As large language models like Vicuna~\citep{zheng2023judging} and Llama2~\citep{touvron2023llama} were trained over documents collected in the real world without sufficient supervision}, such documents might contain noisy information, which could affect the ability of these models during in-context inference. Moreover, recent works~\citep{wei2023jailbreak,pawelczyk2023incontext} also showed that the output of ICL might be significantly manipulated by the label of demonstrations in the prompts, uncovering safety concerns about this prompt form. Motivated by these observations, in this work, we attempt to characterize and understand such ICL ability of transformers \textbf{with noisy labels}. Noisy label learning~\citep{natarajan2013learning} has a wide breadth of literature in modern machine learning research~\citep{xiao2015learning,wang2018iterative,wu2023noisywikihow}, as dataset collection may be costly and noisy~\citep{5206848}. Further discussion on related work can be found in Appendix~\ref{related work}.
{Therefore, the noisy label scenario in terms of ICL, is more practical and deserves a good understanding for advancing the safety and alignment of large language models, as discussed above. }

Given the complexity of language distributions and tasks, directly formulating and analyzing ICL's robustness under noisy label conditions is challenging. To address this challenge, a series of work~\citep{garg2023transformers,bhattamishra2024understanding} suggest studying transformers' ICL capability via \textbf{simple function classes}. Specifically, they propose to assess the ICL ability of Transformers by learning a simple function class in context. Following this research paradigm, our work also investigates the noisy label robustness of ICL using simple function classes, offering a more interpretable and straightforward framework.

In this work, we conduct a comprehensive evaluation of the robustness against noisy labels of ICL and also explore how label noise that exists in the training phase can affect it. Our findings suggest that transformers are indeed fairly robust to label noises in the in-context demonstrations, and it's also possible to further enhance such robustness by adding noises in the training set as data augmentation. Our fruitful and comprehensive experiments provide interesting findings and conclusions for understanding the robustness of in-context learning with noisy labels.

\section{Preliminaries}
\label{pre}
Following the research paradigm of studying the ICL ability of Transformers through a set of simple function classes~\citep{garg2023transformers,bhattamishra2024understanding}, we explore the Transformers' ICL robustness against noisy labels. In this section, we take the \textbf{noisy linear regression} task as an example to demonstrate the overall training-inference pipeline, and other function classes evaluated can also be similarly formulated. The overall pipeline involves the following two parts.

\noindent\textbf{Transformer training with in-context demonstrations.}
In this part, we build a synthetic dataset to train a transformer, aiming for it to acquire ICL capability to learn linear regression functions from in-context demonstrations. Specifically, we consider the class of linear functions
$
    F = \{f_w|f_w(x)=w^Tx, w\in\mathbb R^d\},
$
and each training sample in the dataset can be formulated as 
$
    P=(x_1,f_w(x_1), \cdots, x_k,f_w(x_k)),
$
with input samples and function parameter $w$ both drawn from an isotropic Gaussian distribution, \textit{i.e.} $x_i, w \sim N(0, I_d)$. For noisy-label learning, Gaussian noise $\epsilon_i \sim N(0,\sigma_{train}^2)$ is added to the demonstrations, resulting in $f_w(x_i) + \epsilon_i$. 

\noindent\textbf{Transformer inference with in-context demonstrations.}
After training the transformer model $M_\theta(\cdot)$ with parameters $\theta$, ICL inference is conducted by prompting 
$
    P=(x_1,f_w(x_1), \cdots, x_k,f_w(x_k), x_{q}),
$
where ${(x_i, f_w(x_i)}$ represent \textbf{in-context demonstrations} and $x_q$ is the \textbf{query input}. For in-context inference, a ground-truth function $f_w$ from $F$ is randomly sampled, with $w$ and $x_i$ both drawn from $N(0,I_d)$, resulting in $f_w(x) = w^Tx$. In the noisy-linear regression, noise $\epsilon_i \sim N(0,\sigma_{test}^2)$ is added independently to each demonstration $f_w(x_i) + \epsilon_i$. However, during evaluation, the ground truth of the query $x_q$ is kept as $w^Tx_q$ without adding noise.

\section{Comprehensive evaluation of noisy ICL inference}
\label{sec: eval}
In this section, we focus on the inference stage of ICL with various settings. In our main evaluation, we examine the robustness of the Transformer model with
different distributions of label noises in the linear regression task, and conduct further investigations on the \textbf{non-\textit{i.i.d.} noise setting} (in Appendix~\ref{sec Non iid}), \textbf{different function complexity} (in Appendix~\ref{sec function}), and \textbf{different input dimensions} (in Appendix~\ref{sub set 4 dim}). Following \citep{garg2023transformers}, we construct a dataset with 10,000 prompts, each containing 100 demonstrations with default dimension $d=20$, and train GPT-2 models~\citep{radford2019language} consisting of 12 layers and 8 attention heads and a 256-dimensional embedding for evaluation. All results are averaged over 5 independent experiments.

\noindent\textbf{Noise Distributions.}
In our experiments, we consider the following label noise distributions: 
(1) \textit{Gaussian distribution}: $w^T x_i + \epsilon_i$, where $\epsilon_i \sim N(0, \sigma^2)$;
(2) \textit{Uniform distribution}: $w^T x_i + \epsilon_i$, where $\epsilon_i \sim U( -\sqrt{3\sigma}, \sqrt{3\sigma})$;
(3) \textit{Exponential distribution}: $w^T x_i + {\epsilon_i}$, where $\epsilon_i \sim Exp(\frac{1}{\sigma})$;
(4) \textit{Poisson distribution}: $w^T x_i + {\epsilon_i}$, where $\epsilon_i \sim P(\sigma^2)$;
(5) \textit{Multiplicative distribution}: $w^T x_i \times (1 + \epsilon_i)$, where $\epsilon_i \sim N(0, {\sigma^2})$.
(6) \textit{Salt\&Pepper distribution}: $w^T x_i + \epsilon_i$, where $\epsilon_i$ is either $\text{salt value}$ or $\text{pepper value}$ with probabilities $p/2$ each, and $\sigma = \sqrt{2} p$.

\noindent\textbf{Robustness evaluation metrics.} To evaluate the robustness of the transformer against label noises, following~\citep{garg2023transformers} we compare its performance with a few simple baselines including 
(a) \textit{the least squares estimator}, which computes the minimum-norm linear fit to the in-context examples $(x_i , y_i)$;
(b) \textit{$K$-Nearest Neighbors}, involving the averaging of $y_i$ values for the $n$ nearest neighbors of $x_{query}$;
(c) \textit{the average of the values} $y_ix_i$ to estimate $w$ and calculate the inner product of this estimate with $x_{query}$. 
Note that the least squares offer an optimal estimator for this problem and a lower bound for the best achievable error, while the other two baselines offer a consistent, computationally simpler estimator. Furthermore, we consider the normalized squared error 
$$
    L_P = \frac{(M(P) - w^Tx_{query})^2}{d},
$$
where $P$ is the input prompt and $M$ is the trained model. Considering the presence of label noise, {we define the model achieves the \emph{satisfactory accuracy}} if $L_P \leq 0.5$, thereby establishing a threshold for the noise level under which the Transformer model maintains robustness and providing a criterion to compare between ICL and the simple baselines.

\begin{figure*}[h]
\centering

\begin{tabular}{cccc}
\includegraphics[width=0.3\linewidth]{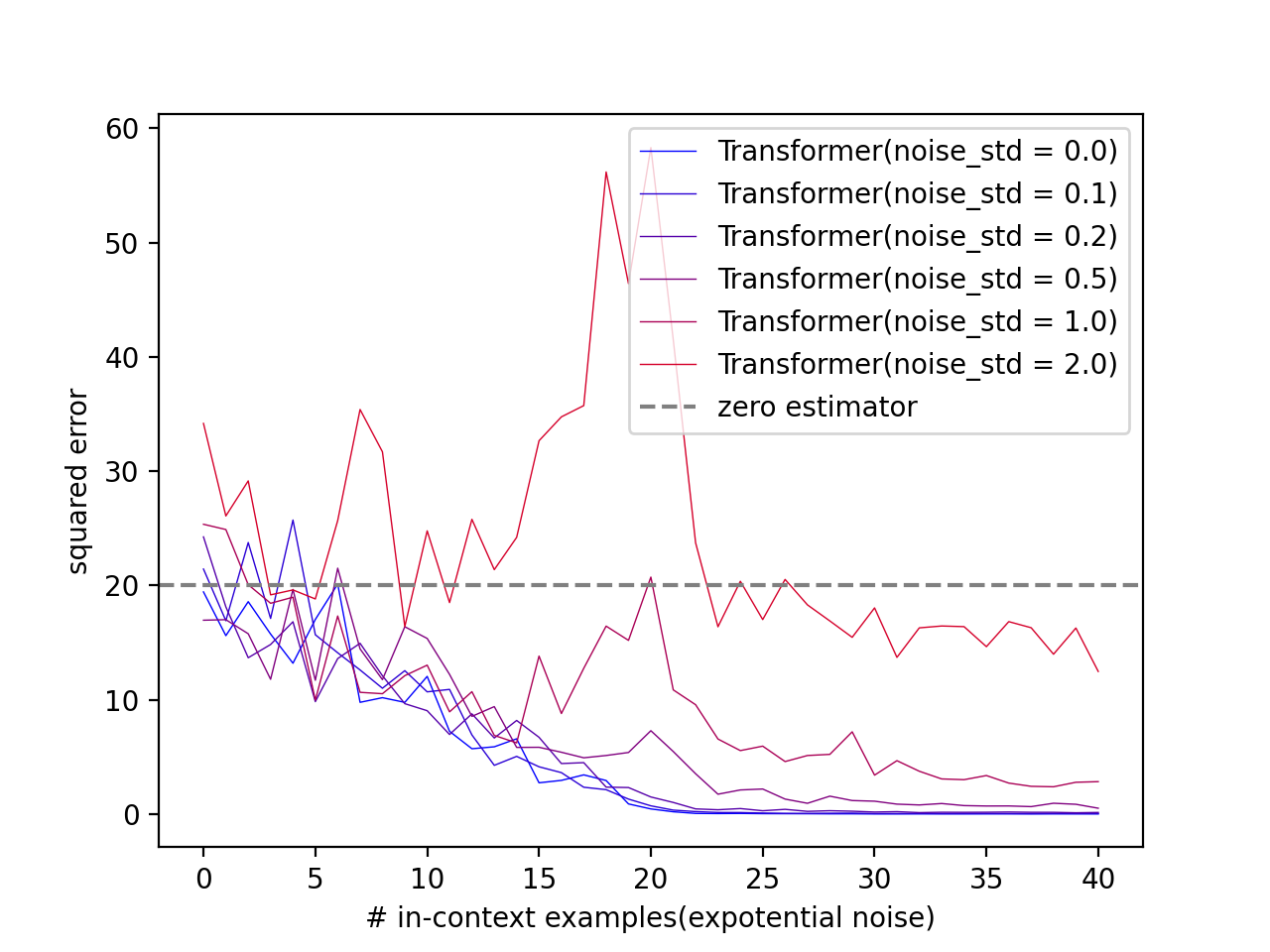} &
\includegraphics[width=0.3\linewidth]{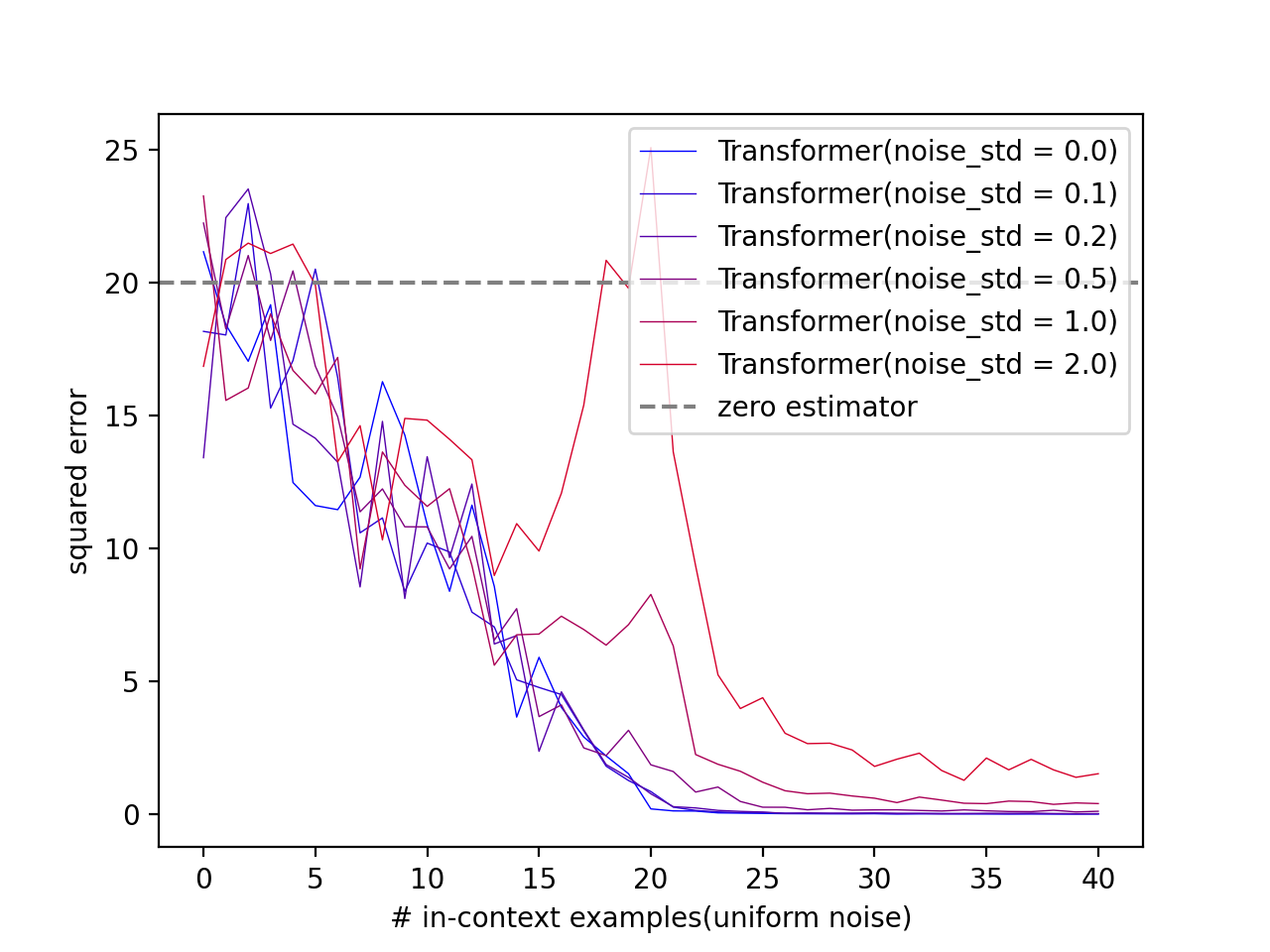} &
\includegraphics[width=0.3\linewidth]{eval_figs/noise_type/diff_std/error_curve_std_expotential.png} &\\
 (a) Gaussian & (b) Uniform & (c) Exponential \\
 \includegraphics[width=0.3\linewidth]{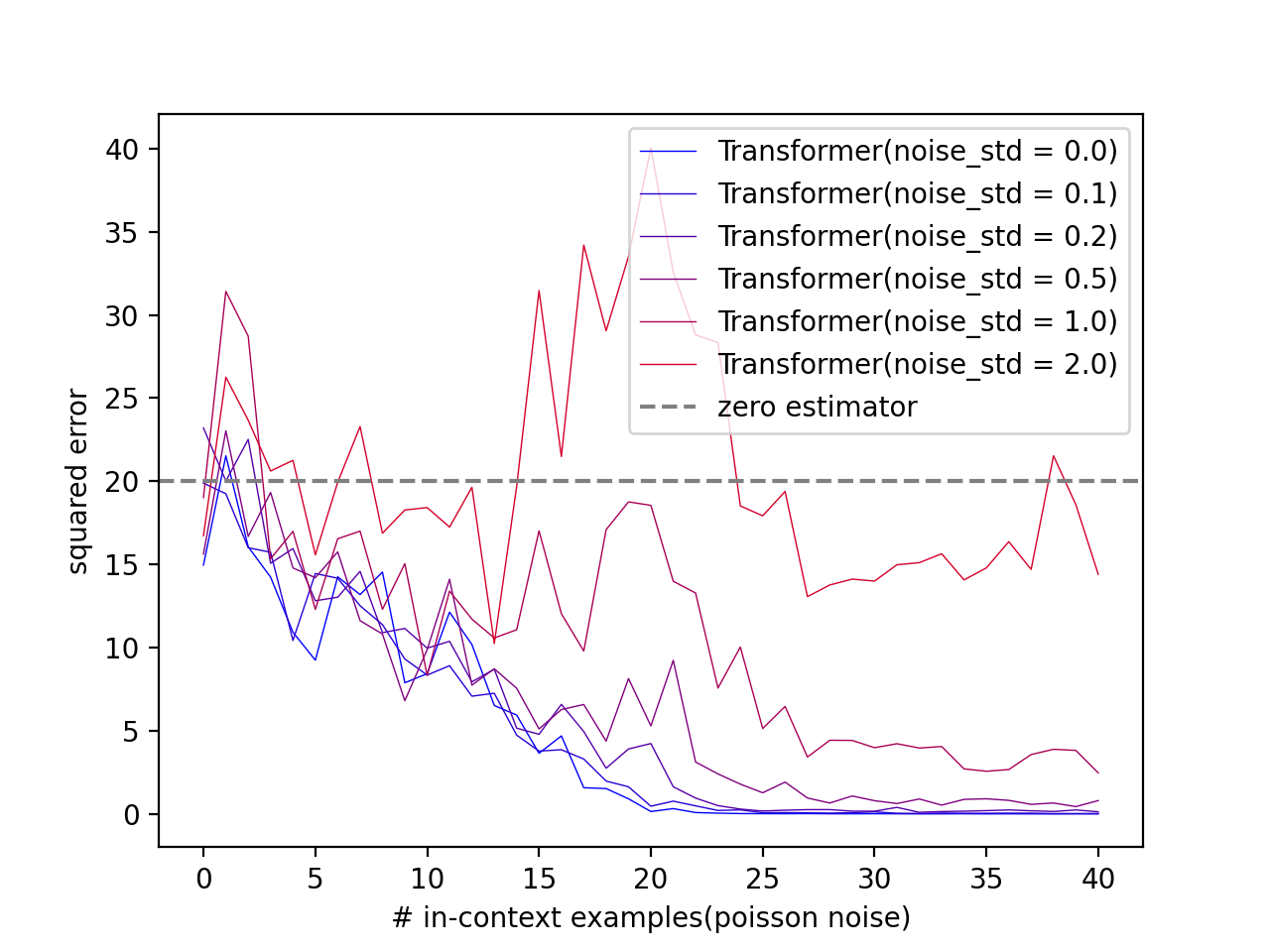} &
\includegraphics[width=0.3\linewidth]{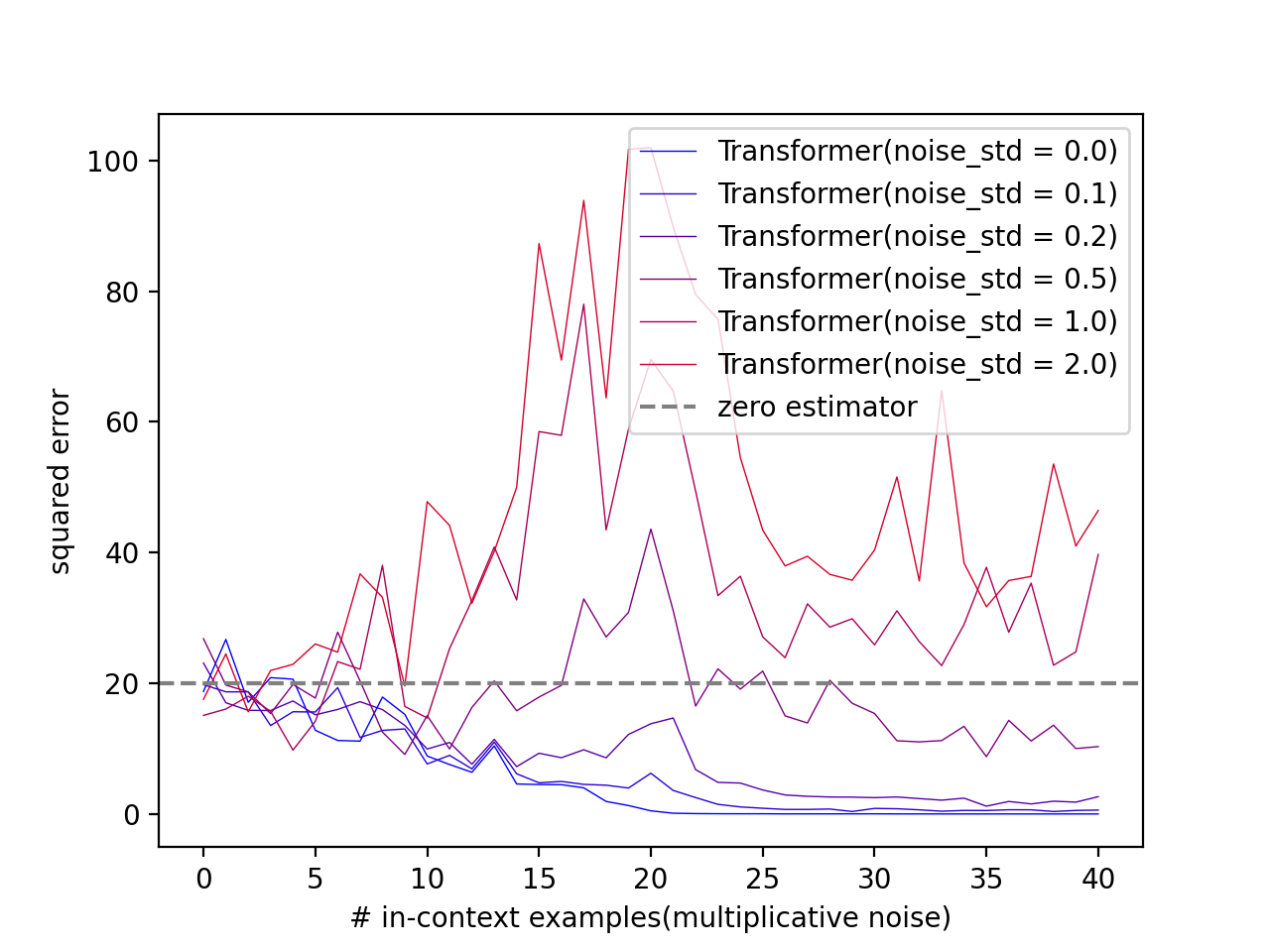} &
\includegraphics[width=0.3\linewidth]{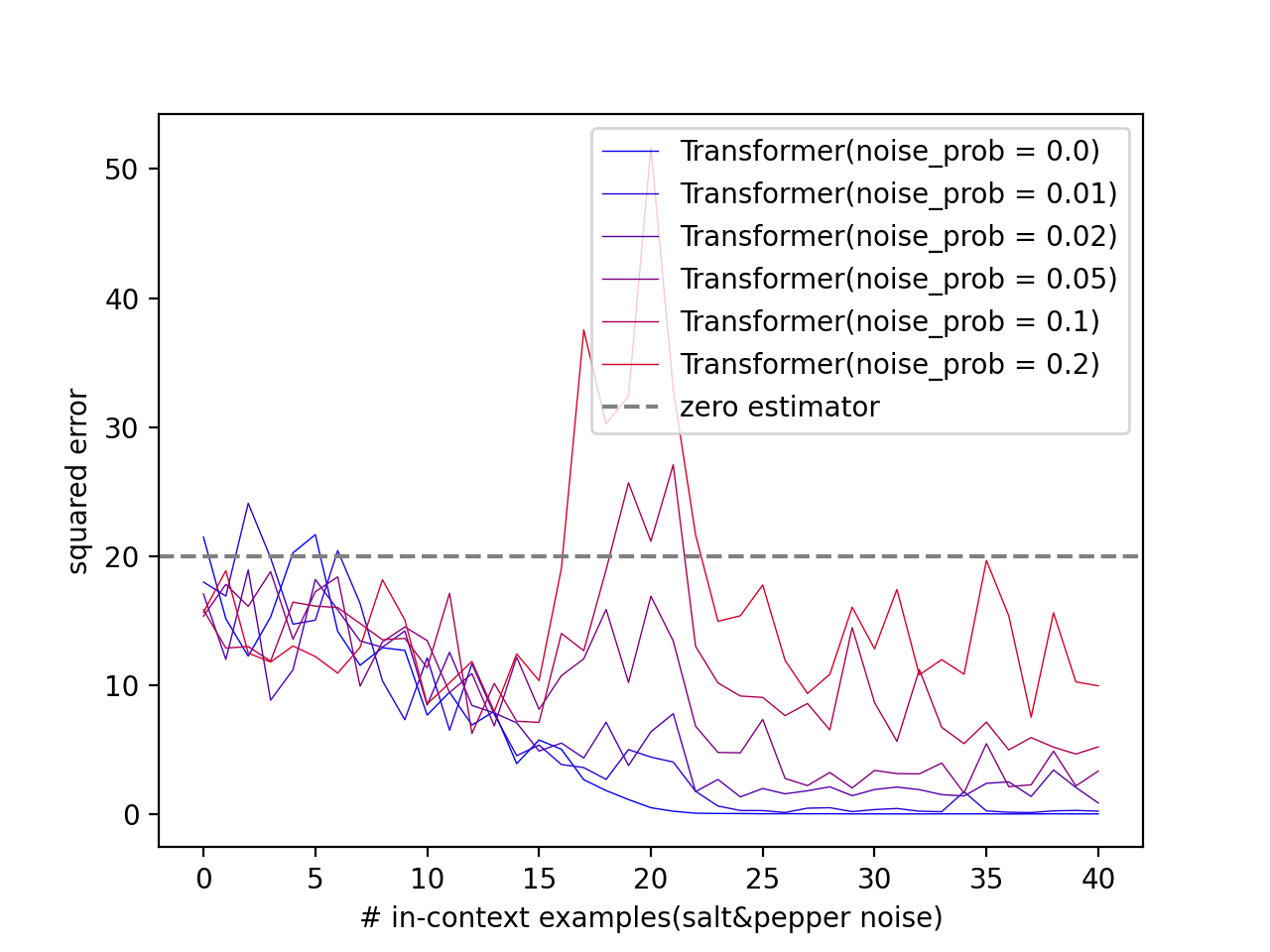} & \\
 (d) Poisson & (e) Multiplicative & (f) S\&P \\
 
\end{tabular}
\caption{Robustness Comparison under Different Noise Types and magnitudes. Each figure represents a noise and each line represents a $\sigma_{test}$. {The X-axis represents the number of in-context examples.} 
}
\label{fig: 5_2}
\end{figure*}

\noindent\textbf{Robustness analysis.}
As discussed above, we employ the loss functions introduced above to evaluate the trained Transformer models and the baselines across different noise types and varying levels of label noise. 
We plot the comparison of the transformer and other baselines under these settings in Figure~\ref{fig: 5_1} in Appendix~\ref{app_A2}, including $\sigma_{test}\in\{0.2,0.5, 1.0\}$, and also compare the robustness of the Transformers against different noise magnitude $\sigma_{test}$ of a specific noise type in Figure~\ref{fig: 5_2}.

From these results, we can directly draw the following observations:

\textbf{i) Inadequate in-context examples.} When the number of in-context examples falls below the input dimension ($d=20$), the model's loss closely aligns with that of the least squares estimator.

\textbf{ii) Near input-dimension number of examples.} When the number of in-context examples nears the input dimension ($d=20$), significant errors arise in the least-square estimator, whereas the ICL model's accuracy continues to improve.

\textbf{iii) {Sufficient} in-context examples.} Once the number of in-context examples exceeds $d=20$, the ICL model's performance rapidly and significantly improves.

Furthermore, we can find that under \textbf{symmetric} noises (Gaussian, Uniform, Exponential, Poisson), transformers exhibit better performance in comparison to other baselines. Besides, when confronted with higher noise levels, the ICL model exhibits a challenge in ignoring label noise. This observation aligns with intuition, as inference in noisy linear regression mirrors the least squares solution with appropriately structured $L_2$ regularization. 
Additionally, upon introducing \textbf{Multiplicative} and \textbf{Salt\&Pepper} noise, we observe that the transformer model does not perform well (worse than baselines and zero estimators). This discrepancy suggests that transformer models might not generalize well to noise that deviates significantly from the \textbf{symmetric or discrete distributions} typically encountered in natural data.

Finally, we also find that there may exist a distinct threshold for each noise type, beyond which the transformer model's performance cannot outperform baselines. Once the noise level, denoted as $\sigma$, surpasses this threshold, the noise perceptibly impacts the model, rendering it non-negligible in its influence on performance. Through extensive evaluations, we can estimate such thresholds for different noise distributions and summarize them in Table~\ref{table:5_2}.

\vspace{-10pt}
\begin{table}[h]
\caption{Estimated robustness threshold of label noise $\sigma$ to perform comparably with the case of no noise.}
\centering
\begin{tabular}{c|c|c|c|c|c|c}
\toprule
\textbf{Distribution} & \textbf{Gaussian} & \textbf{Uniform} & \textbf{Exponential} & \textbf{Poisson} & \textbf{Multiplicative} &  \textbf{S\&P}\\
\midrule \textbf{Threshold $\sigma$}      & 0.45     & 1.10    & 0.39        & 0.43 & 0.23 & 0.22 \\
\bottomrule
\end{tabular}
\label{table:5_2}
\end{table}

\section{Influence of training with noisy labels on robustness}
\label{sec: train}

In this section, we further explore how adding label noises in the training dataset influences the robustness of the Transformer during in-context inference. Since the real-world training corpus for large-scale Transformers (\textit{e.g.}, LLMs) may be noisy and lack supervision, we attempt to seek insights into whether such noises may be unexpectedly helpful in improving the robustness of the model during ICL by this evaluation framework. Intriguingly, we find that adding such noises to the training labels can indeed enhance the desired robustness.

\noindent\textbf{Experiment Set-up.}
To construct a noisy training set, instead of vanilla demonstrations $\{x_i, f_w(x_i)\}$, we add Gaussian noises to the labels as $\{x_i, f_w(x_i) + \epsilon_i\}$ with $\epsilon_i\sim N(0,\sigma^2)$. We still adopt the GPT-2~\citep{radford2019language} model architecture and compare $\sigma_{\text{train}} \in \{0.2, 0.4, 0.6,0.8, 1\}$. In this experiment, we further take the model size into consideration and compare the following three model sizes:
\emph{standard} (8 heads, 12 layers, embedding size = 256, which is adopted in the experiments above), \emph{small} (4 heads, 6 layers, embedding size = 128), and \emph{tiny} (2 heads, 3 layers, embedding size = 64). 

\begin{figure}[h]
\centering
\includegraphics[width=0.7\linewidth]{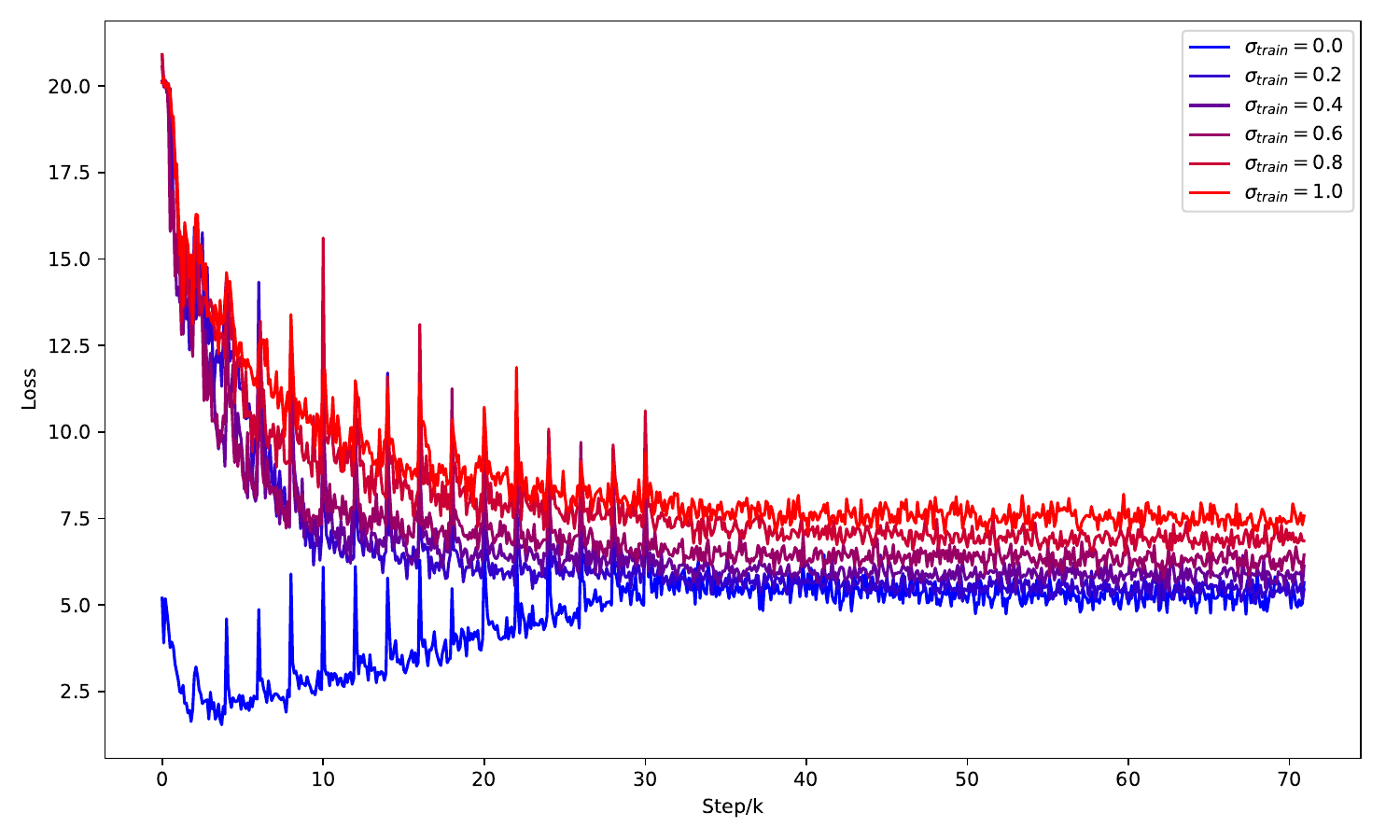}
\vspace{-10pt}
\caption{The training loss for a standard-sized transformer with Gaussian noise of $\sigma_{train} \in \left \{ 0.0,0.2,0.4,0.6,0.8,1.0 \right \}$.}
\label{fig:training_loss}
\end{figure}

\noindent\textbf{Training convergence with noisy labels.} 
For each $\sigma$, we train the model with corresponding label noise distribution and plot the training loss curve in Figure~\ref{fig:training_loss} with each line representing a model. 

Similar to~\citep{garg2023transformers}, we used curriculum learning during training where we started from a low-dimensional space and gradually increased to 20. Note that the model associated with $\sigma_{train}=0.0$ represents the case of no-label noise, which is exactly the standard training setting. By inspecting the convergence situation, it is clear that for small label noises ($\sigma<0.6$), the training is well converged and the final loss is very close to the standard training. Even in the case of large noise variance ($\sigma=0.6, 0.8, 1$), the training can still converge to a certain extent. The main difference between training with noise and without noise is at the beginning where the loss remains high even with the adoption of curriculum learning. However, curriculum learning still works well since all the models nearly converge at step 30k.

\noindent\textbf{Robustness analysis with noisy label training.}
{We further conduct in-context inference with different inference Gaussian noise $\sigma_{test}\in \{0.2, 0.4, 0.6, 0.8, 1\}$ (the same as what we used to train our models). Still, we focus on the class of noisy linear functions $F=\{f_w | f_w(x) = w ^Tx +\epsilon, w \in R^d\}$, in $d=20$ dimensions. 
To ensure that the experimental results are convincing and reproducible, we repeated the experiment 1280 times and took an average value to {plot} the figure. {Moreover, for the sake of fairness, we use the same set of prompts for different models in the same setting during inference.} The results are  Figure~\ref{fig:training_with_noise}.

\begin{figure}[htbp]
\centering
\includegraphics[width=\linewidth]{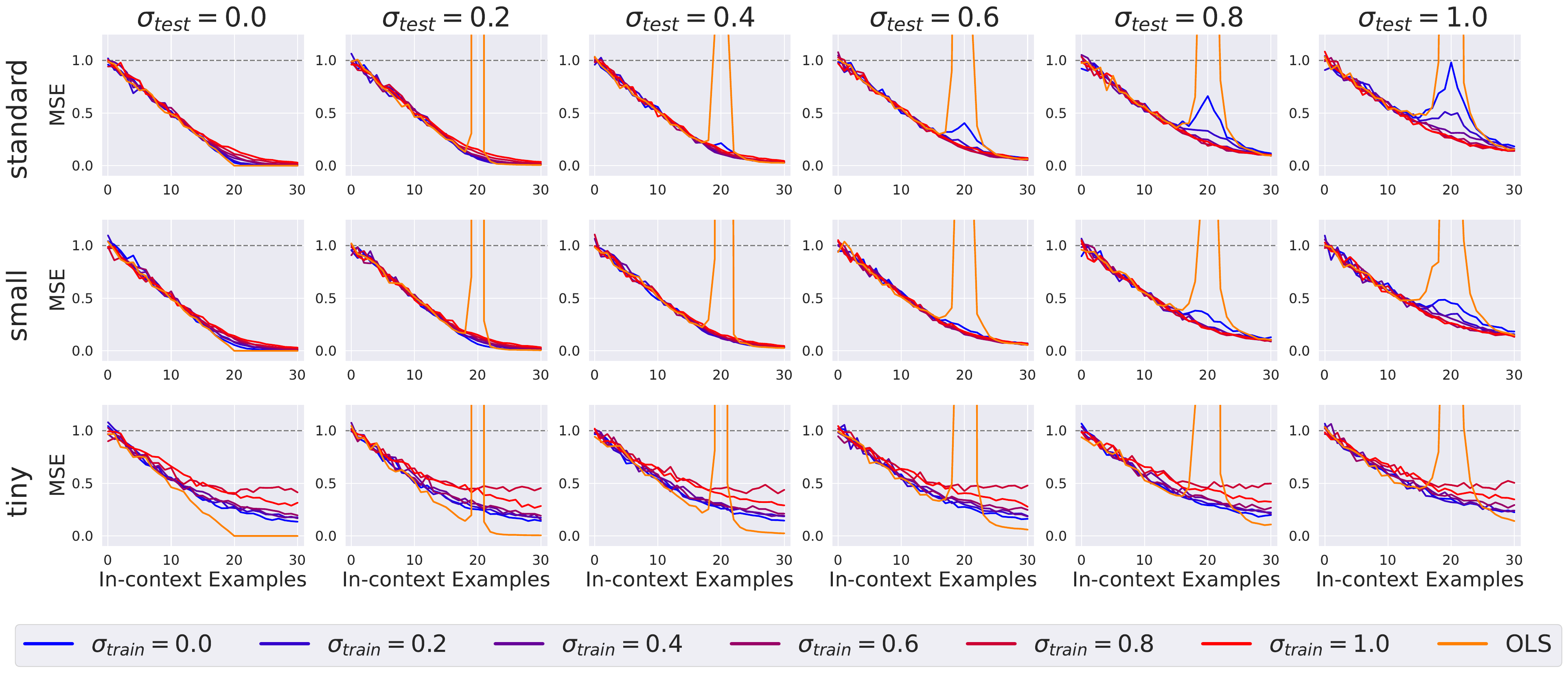}
\vspace{-10pt}
\caption{Effects of Gaussian Noise Magnitude and Model Size on Performance. }
\label{fig:training_with_noise}
\end{figure}

\noindent\textbf{Robustness improved with training label noises} Intriguingly, we find that adding label noises in the training dataset can indeed improve the robustness of Transformers during ICL inference. Specifically, taking the case of standard model size as an example (the first row), though the performance of different models with small test noise $\sigma_{test}<0.5$ is small, when the test noise increases $\sigma_{test}>0.5$, the models trained with noises (red lines) outperforms the models trained without noise and with small noise ($\sigma_{train}<0.2$, blue lines) significantly. Similar phenomena are also observed in smaller model sizes, though the improvement of robustness exhibited in large test noises is relatively decreased.

Besides, we can also observe that for smaller test noises and the natural case ($\sigma_{test}=0$), the performance of models trained with large training noises is similar to the vanilla cases for larger model sizes (standard and small), but decreases notably for tiny model size, showing the importance on the model capacity in learning with noisy labels. The conclusion is consistent with training with LLMs, where training with noisy corpus can still help the model generalize well and be robust against noisy labels during ICL~\citep{min2022rethinking} given the large size of the model.

\section{Conclusion}

\label{sec conclusion}
In this paper, we present a comprehensive study on the robustness of Transformers' in-context learning ability against label noises. First, we conduct extensive experiments covering different noise types, magnitudes, function complexities, and input dimensions to show that Transformers retain their ICL efficiency against various label noises, along with sufficient analysis and discussions. Furthermore, we discover that introducing similar noises into the training set enables good convergence and potentially enhances the robustness of transformer models. Our findings contribute to a deeper understanding of Transformers' ICL adaptability to label noise and provide new insights into ICL research in real-world models.

\section*{Acknowledgement}

This work was sponsored by the Beijing Natural Science Foundation's Undergraduate Initiating Research Program (Grant No. QY23041) and the National Natural Science Foundation of China (Grant No. 62172019).

{
\small
\bibliographystyle{plainnat}
\bibliography{ref}
}

\newpage
\appendix
\section{Additional related work}
\label{related work}
\subsection{Understanding In-context Learning} The mysterious ability of in-context learning~\citep{brown2020language,dong2023survey}, which typically occurs in attention-based model architectures like transformers~\citep{vaswani2023attention}, has attracted significant research interest in understanding its underlying mechanism and designing better learning algorithms~\citep{lu2023emergent,min2022rethinking}. Without modifying model parameters, these models can conduct various downstream tasks with a few input-output pairs as demonstrations included before the test input. 

However, given the complexity of natural languages, a popular research paradigm of understanding the ICL ability of Transformers is characterizing their learnability through a set of simple function classes, like linear functions~\citep{garg2023transformers} and discrete functions~\citep{bhattamishra2024understanding}, which are more interpretable and easier to formulate. There are also other works that investigate such ability of ICL through simple functions and propose various understandings, \textit{e.g.} interprets the inference with in-context demonstrations as implicit gradient decent~\citep{akyürek2023learning,pmlr-v202-von-oswald23a,bai2023transformers,wang2024language}. Specifically, they showed that transformers can learn the specified task through these demonstrations with implicit optimization in the hidden spaces of transformers. Besides, there are also interpretations of ICL through other perspectives, like Bayes inference~\citep{xie2022explanation} and PAC-learnability~\citep{wies2023learnability}.

\subsection{Noisy label learning}
{Modern} deep learning methods commonly face the presence of noisy labels in the training data, since data collection and annotation may be costly and biased~\citep{5206848}. Such noises in labels may be symmetric, asymmetric, and even from the open set that is not contained in the training classes~\citep{cordeiro2020survey}. To tackle this issue, numerous efforts have been made to robustify the training process against such noises. Typical approaches include estimating noise transition matrix~\citep{patrini2017making}, designing robust loss functions~\citep{Manwani2011NoiseTU,wang2019symmetric}, sample weighting~\citep{guo2018curriculumnet,wang2018iterative} and selection~\citep{jiang2018mentornet}.

Though broadly explored in this literature, learning with noisy labels is still an open problem in modern machine learning research~\citep{song2022learning}. Moreover, there are also concurrent threads toward studying the robustness against label noise for the text modular~\citep{Garg_2021,zhu2022bert}. Despite broad explorations, few works have investigated the robustness of in-context learning with noisy labels. 
Though studying the ICL ability of transformers through noisy linear regression, \citep{garg2023transformers} only studies the standard Gaussian distribution noise, which is only one specific noise type and magnitude in our work. In this paper, we take steps further to systematically investigate the robustness of ICL against various label noise settings, including both the train and inference phases, {and various types and magnitudes of noises.}

\subsection{Language model safety and alignment}
With the milestone success of the fast-paced development of large language models (LLMs), concerns regarding their potential for harmful generation and malicious usage have emerged~\citep{bommasani2022opportunities,chen2023combating,liu2023towards,zhang2024towards}, which are typically referred to as the jailbreaking issue~\citep{zou2023universal,wei2023jailbroken,dong2023robust,zhang2024boosting}. Such risks further extend to the in-context learning scenario, as recent work~\citep{wang2023adversarial,wei2023jailbreak} showed, it is possible to manipulate the safety and alignment of language models by maliciously inducing noisy labels in the demonstrations. Therefore, it is of significance to study the robustness of in-context learning with label noises. 

{Moreover, different from defense against conventional adversarial noises~\citep{goodfellow2014explaining,carlini2017evaluating,liu2022practical,chen2023rethinking,wei2023sharpness} among which adversarial training methods~\citep{madry2017towards,zhang2019theoretically,wei2023cfa,zhang2024duality} that adding adversarial noises in the training loops are shown to be the most effective among various defense methods~\citep{carlini2017adversarial,athalye2018obfuscated}, it is recently shown that AT may not be an effective defense~\citep{jain2023baseline} for LLMs. Such difference further underscores the challenge of the interpretability of internal mechanism~\citep{islam2021explainable,rauker2023transparent,zou2023representation} and adversarial examples~\citep{tsipras2018robustness,wang2020unified,li2023adversarial} in the context of LLMs.}  In this work, we explore a similar problem: whether adding label noises in the training demonstrations can enhance the corresponding robustness during inference.

\section{Complete visualizations for main evaluation in Section~\ref{sec: eval}}

\label{app_A2}

\begin{figure}[!h]
\centering

\begin{tabular}{cccc}
    {\small Gaussian}  & \includegraphics[width=0.24\linewidth,valign=c]{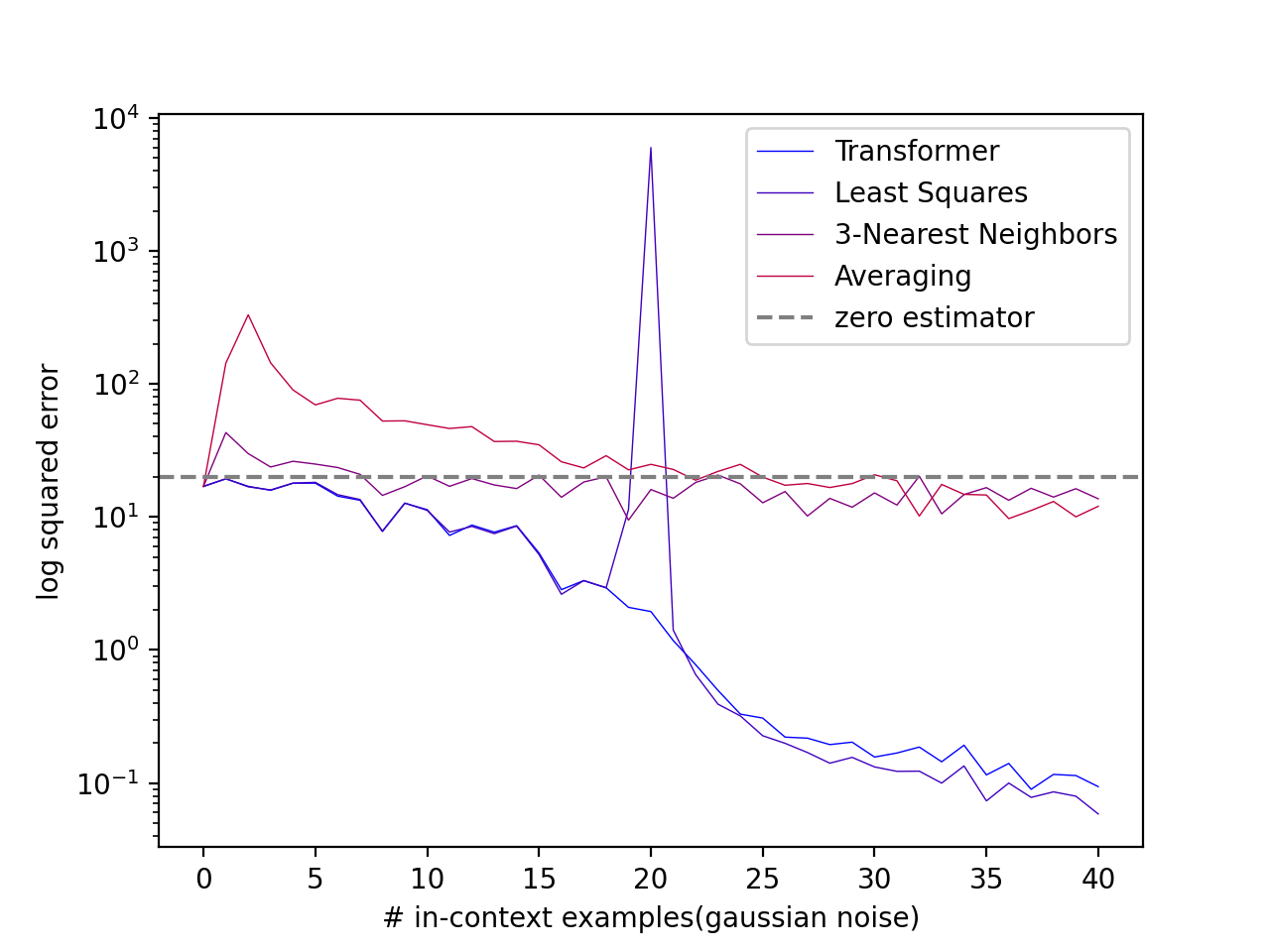} &  \includegraphics[width=0.24\linewidth,valign=c]{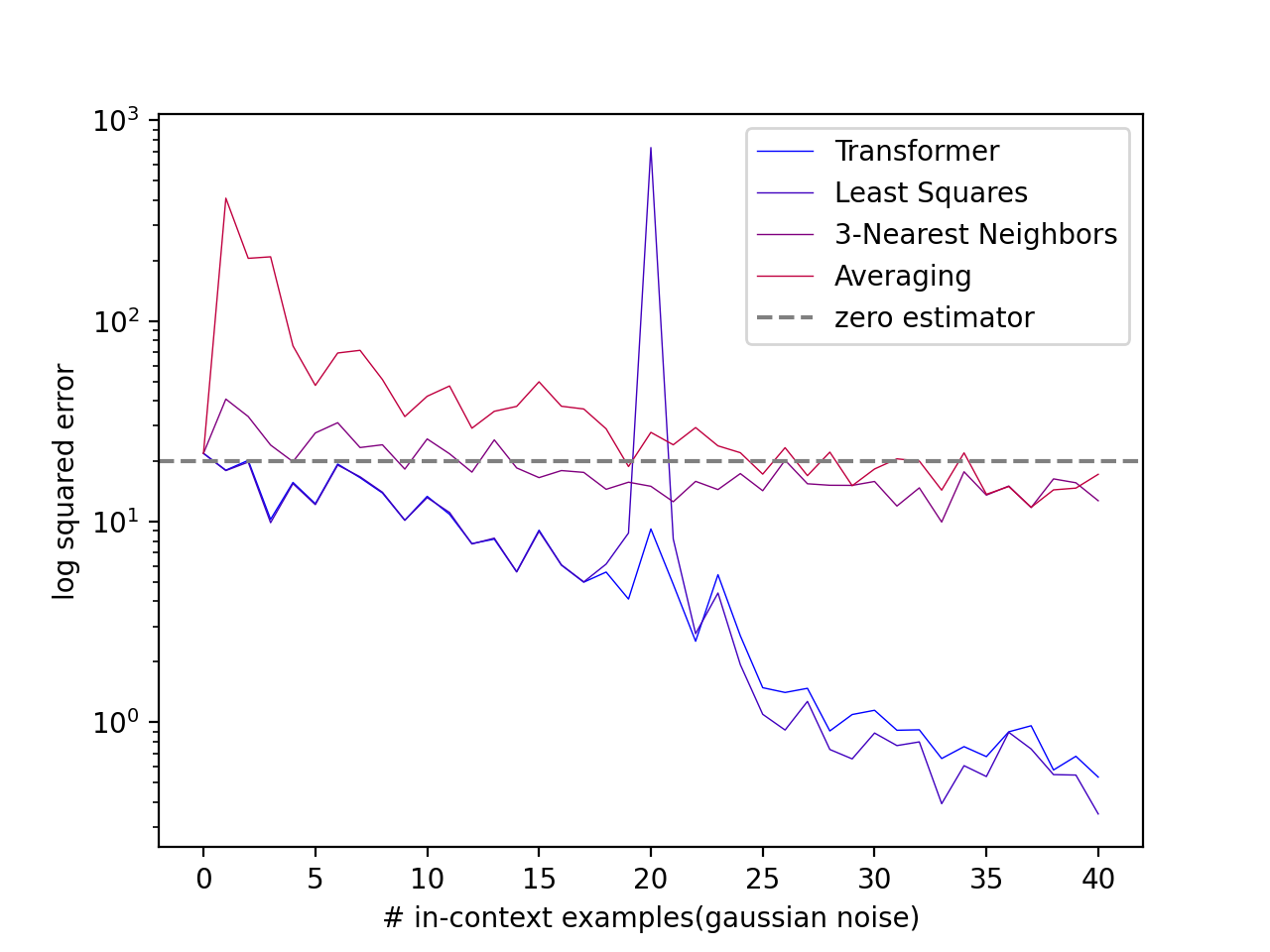} & \includegraphics[width=0.24\linewidth,valign=c]{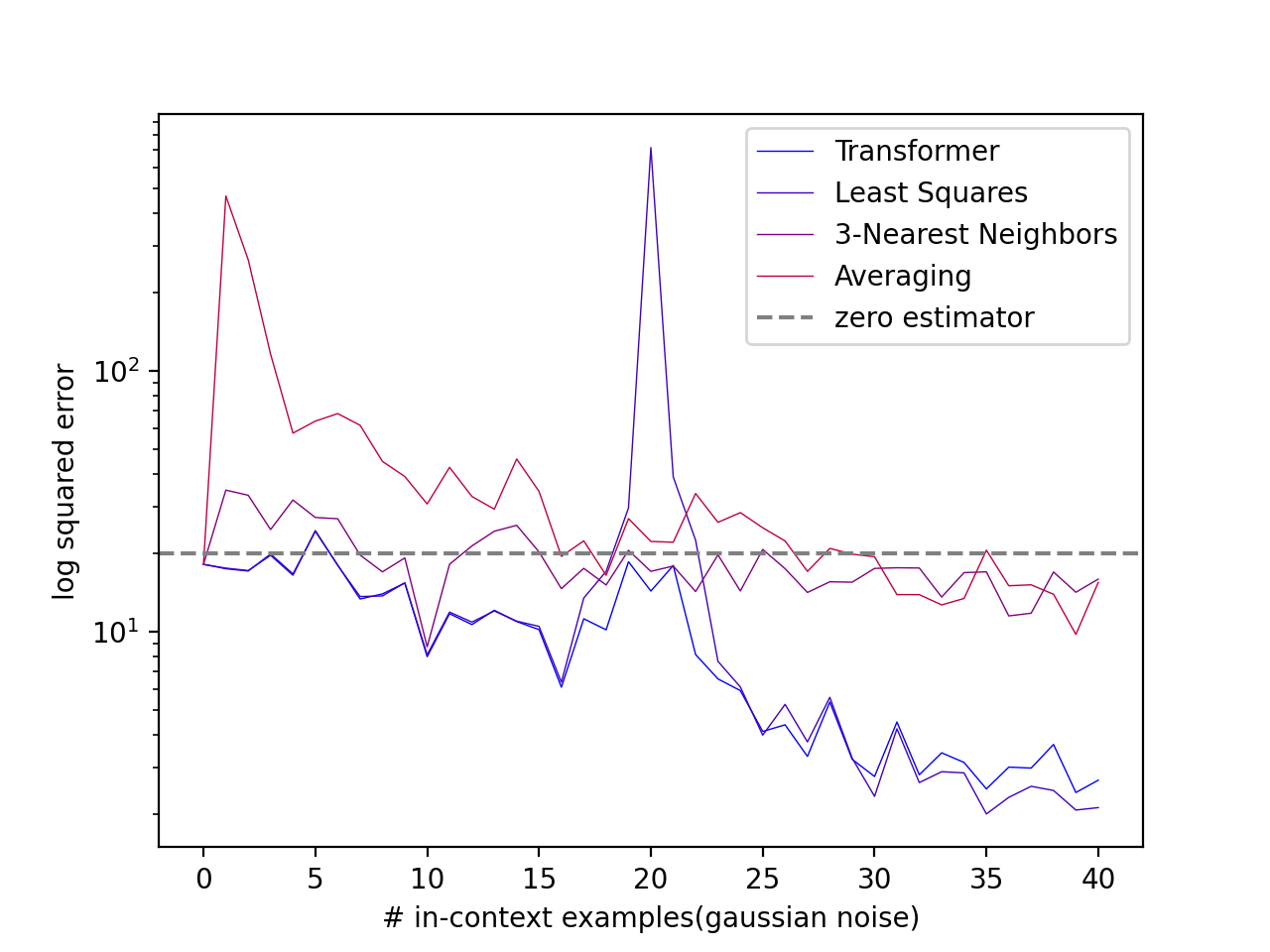} \\
    {\small Uniform} & 
    \includegraphics[width=0.24\linewidth,valign=c]{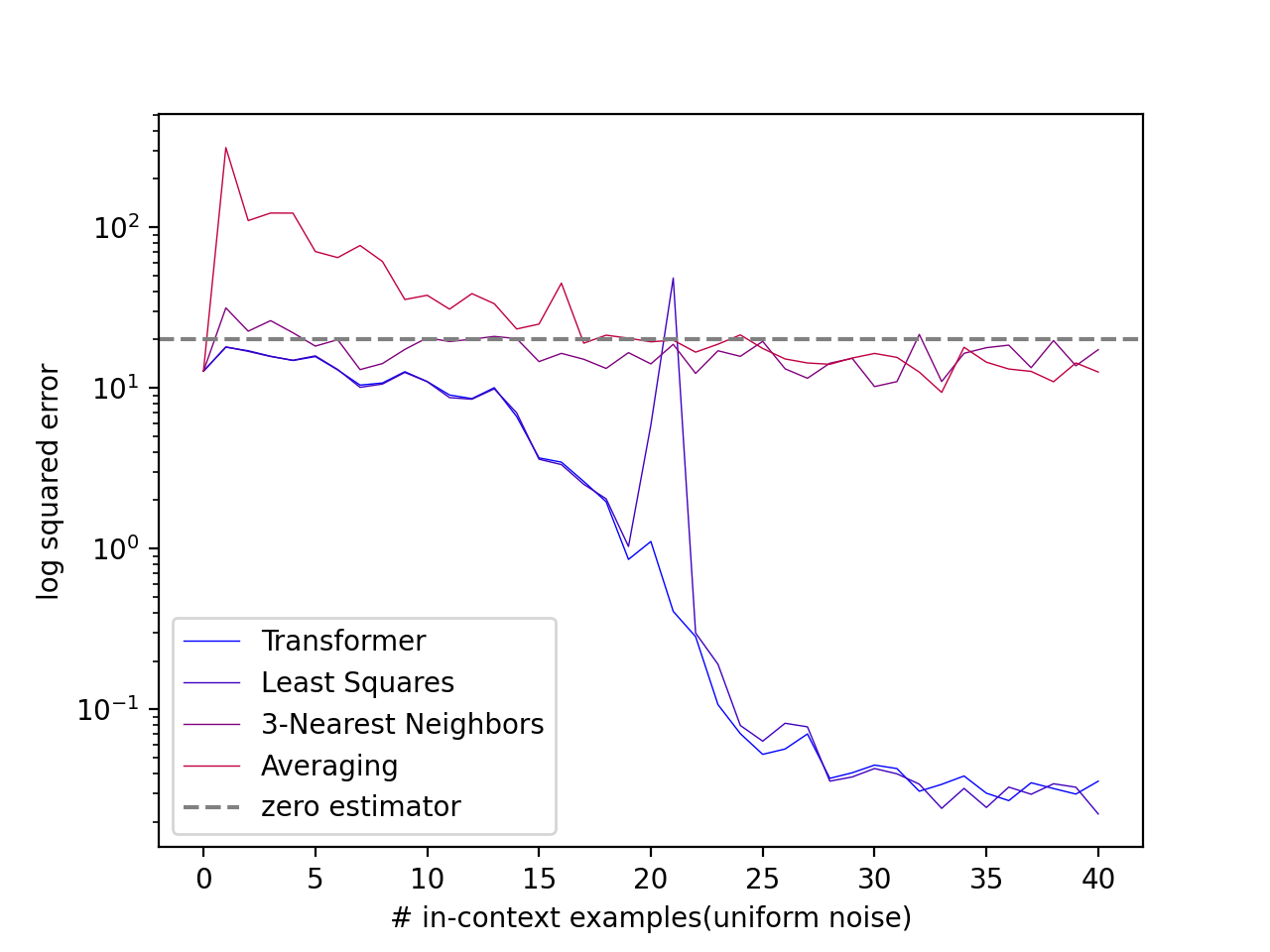} & 
    \includegraphics[width=0.24\linewidth,valign=c]{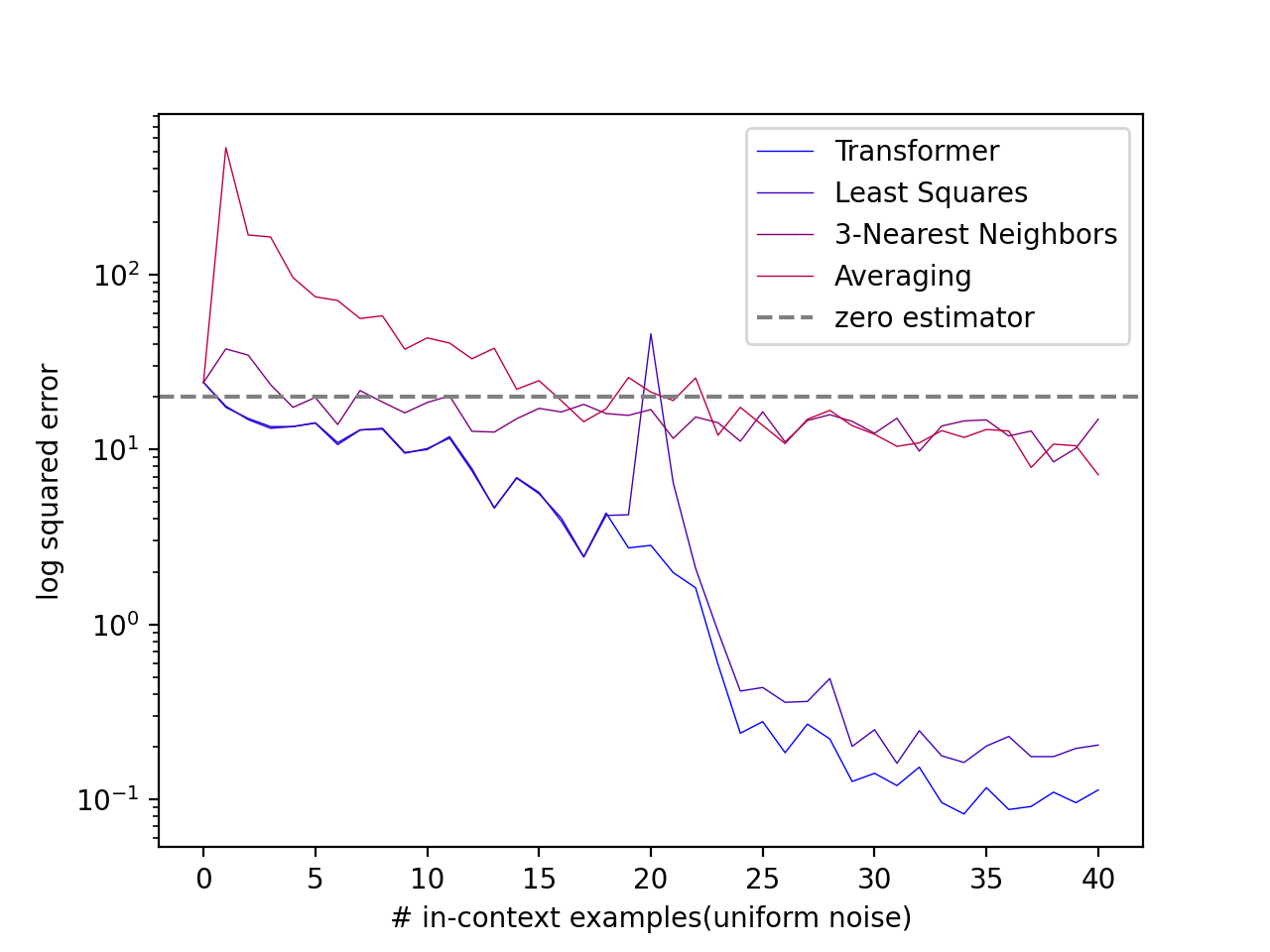} & 
    \includegraphics[width=0.24\linewidth,valign=c]{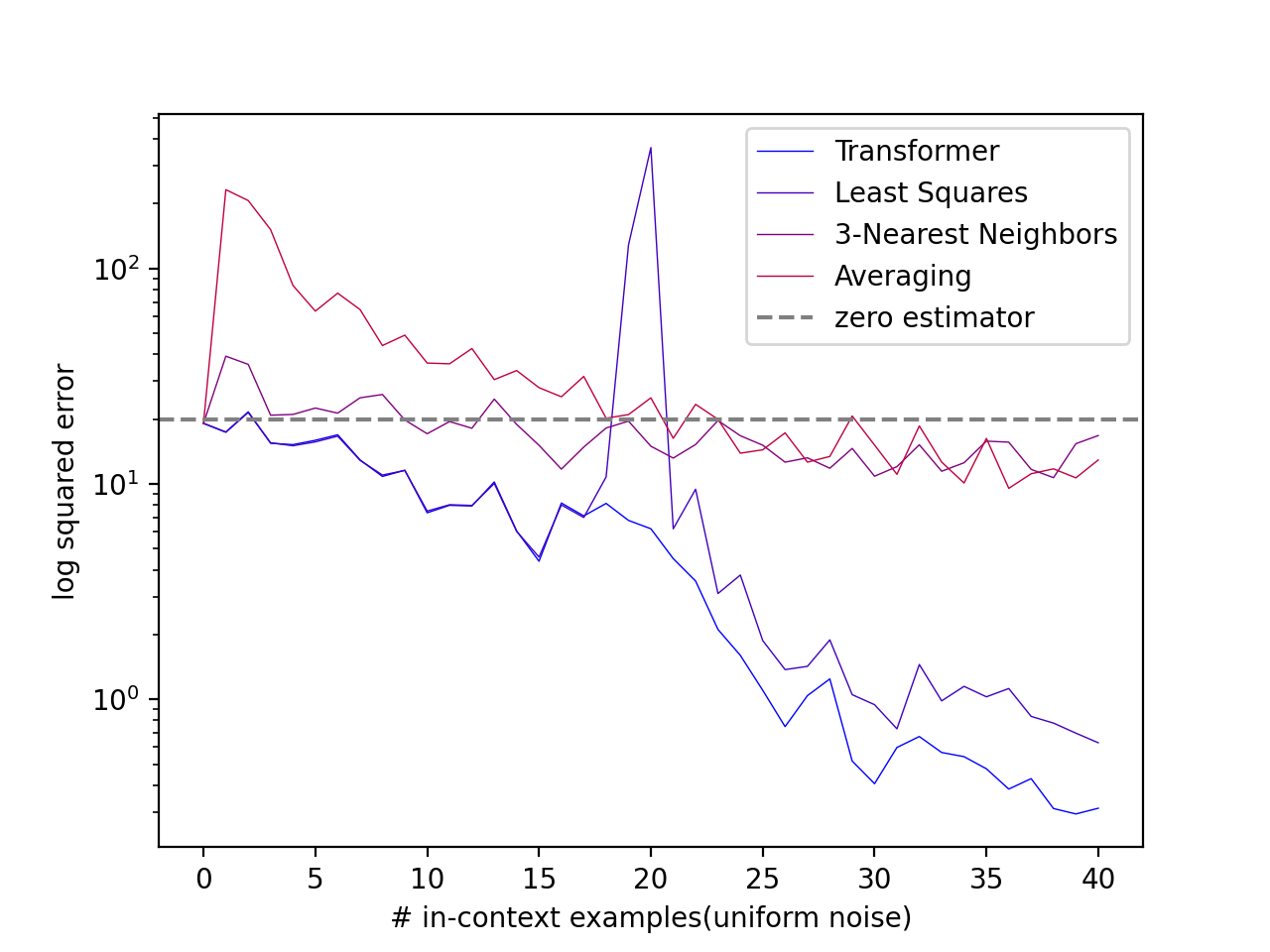} \\
    {\small Exponential} & \includegraphics[width=0.24\linewidth,valign=c]{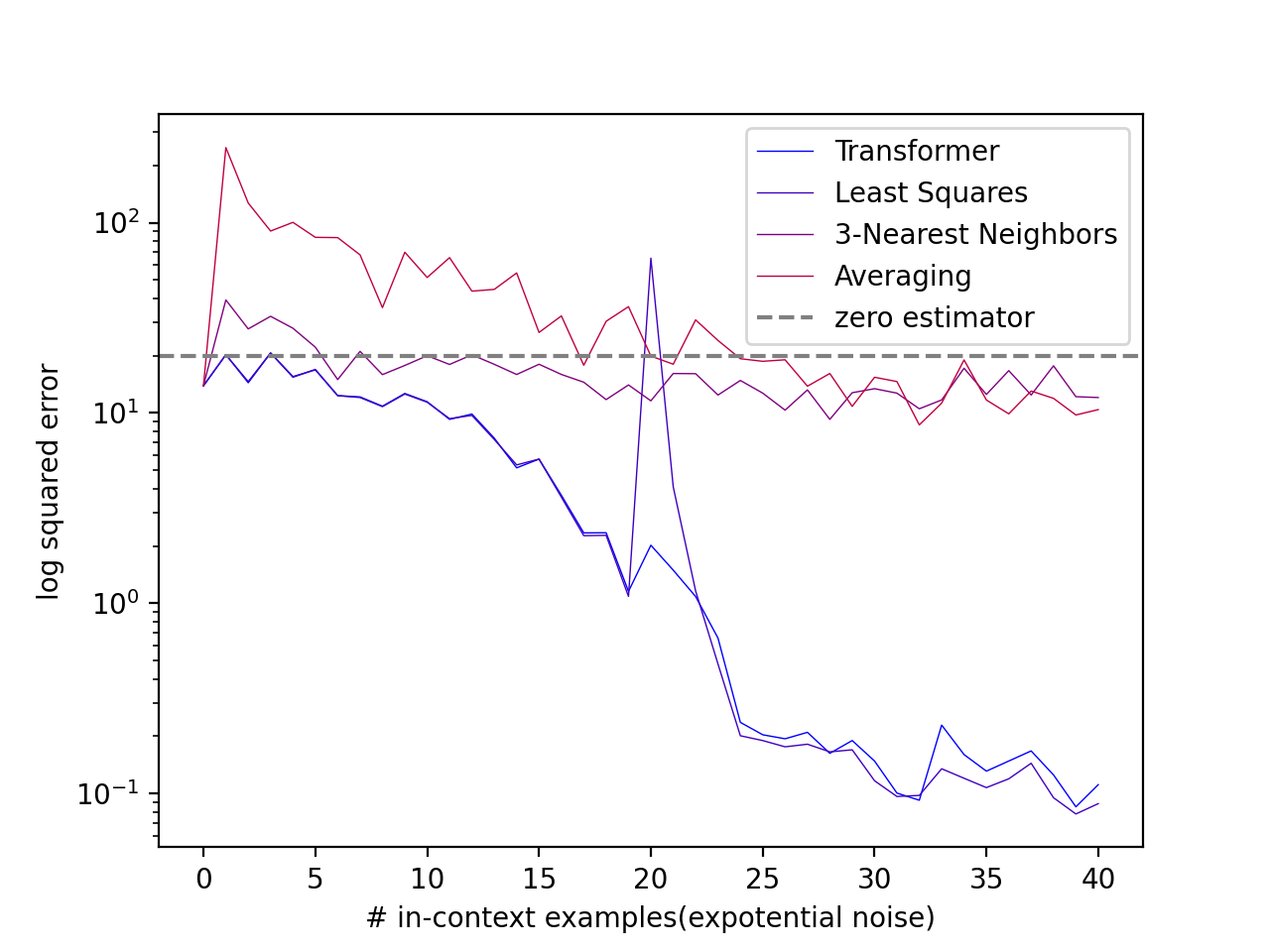} &
    \includegraphics[width=0.24\linewidth,valign=c]{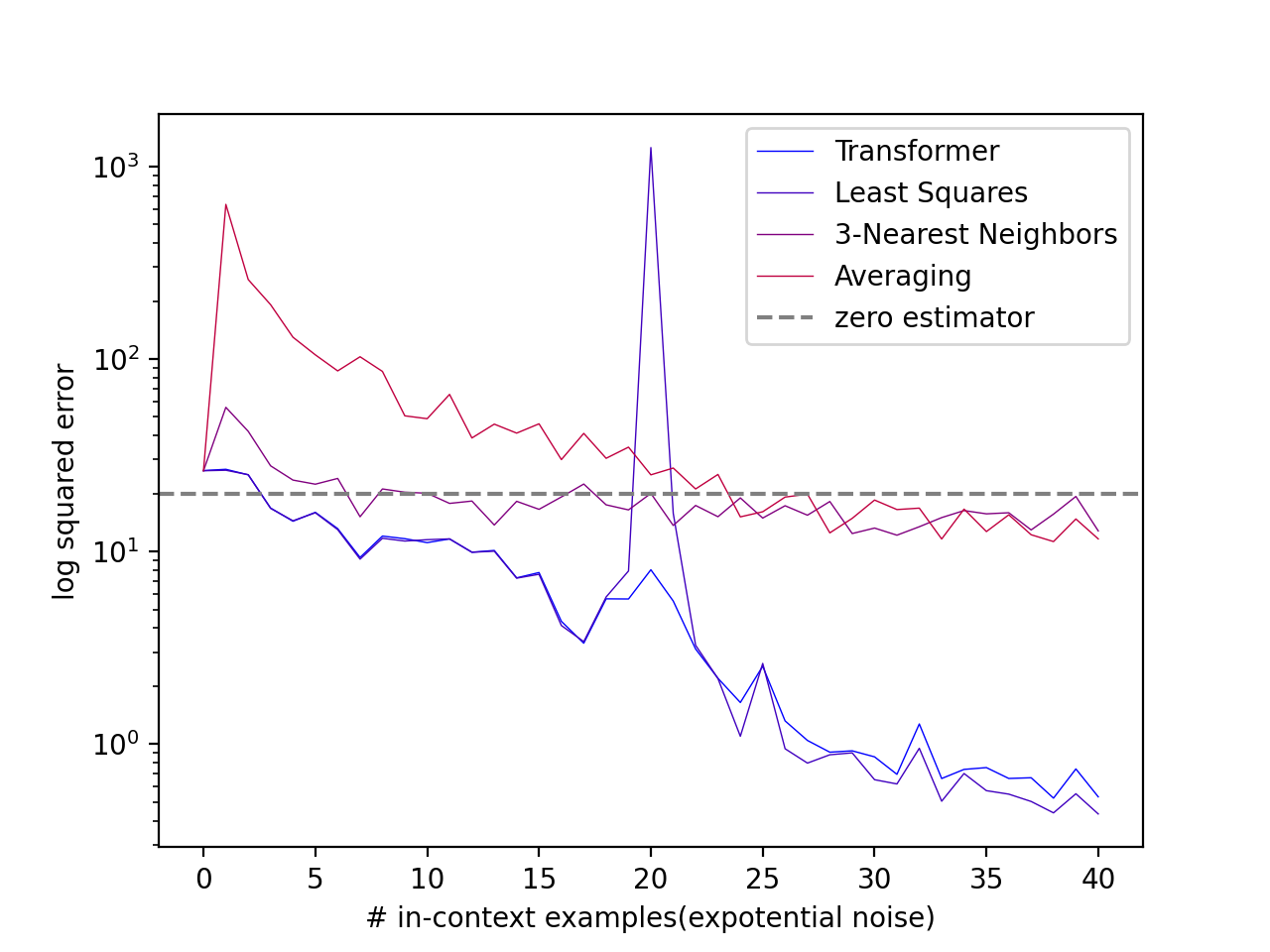} &
    \includegraphics[width=0.24\linewidth,valign=c]{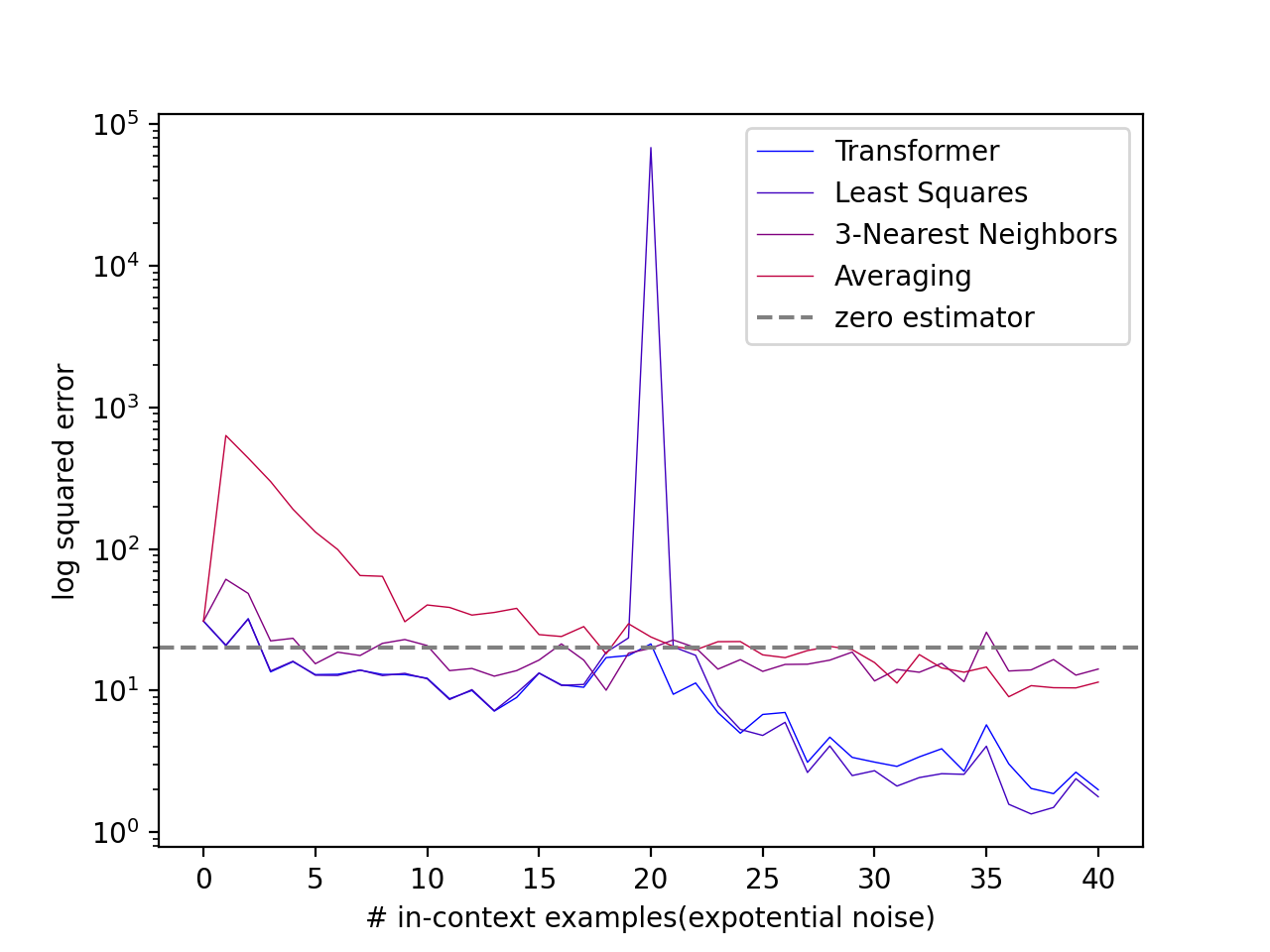} \\
    {\small Poisson} & 
    \includegraphics[width=0.24\linewidth,valign=c]{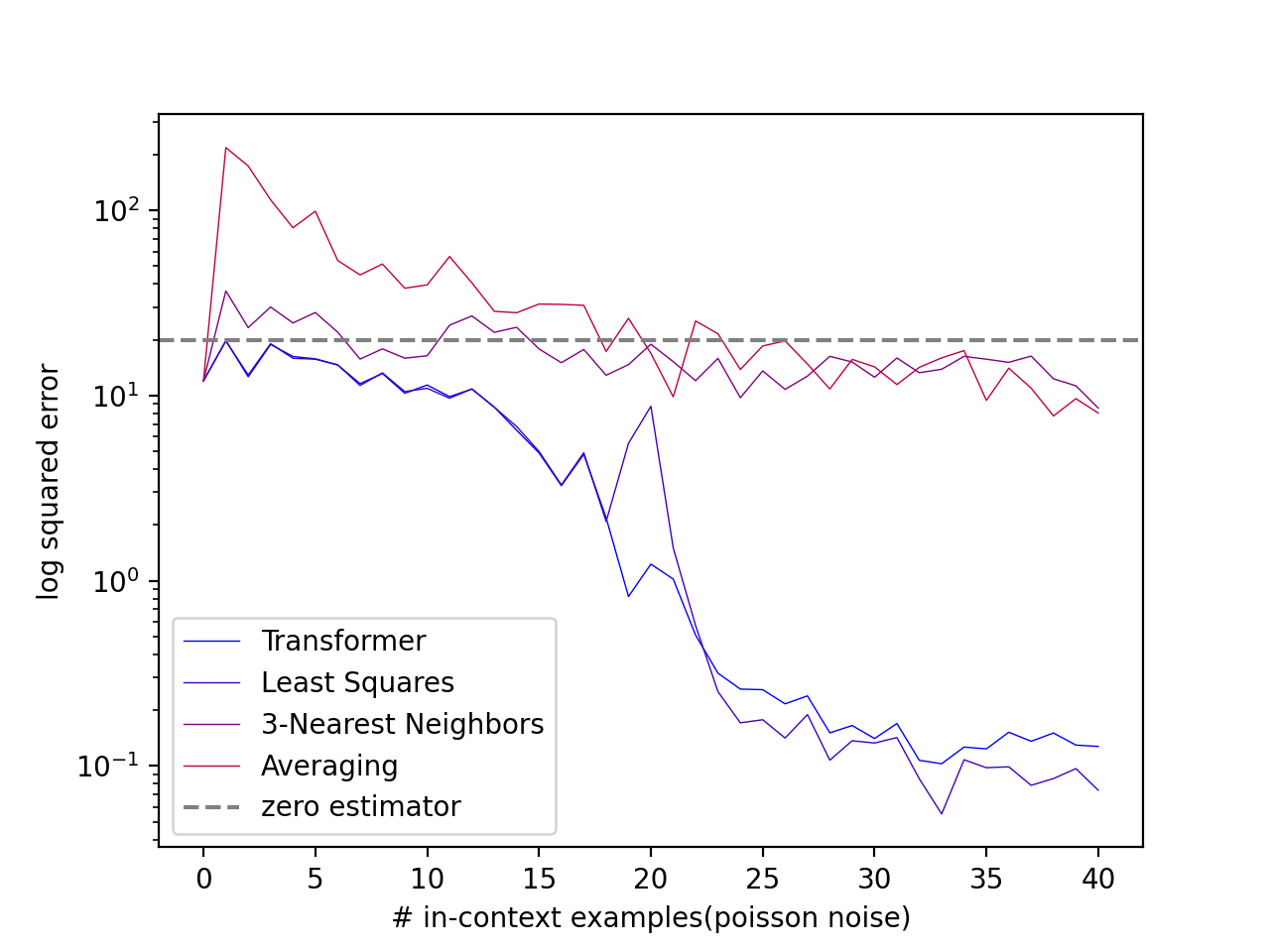} &
    \includegraphics[width=0.24\linewidth,valign=c]{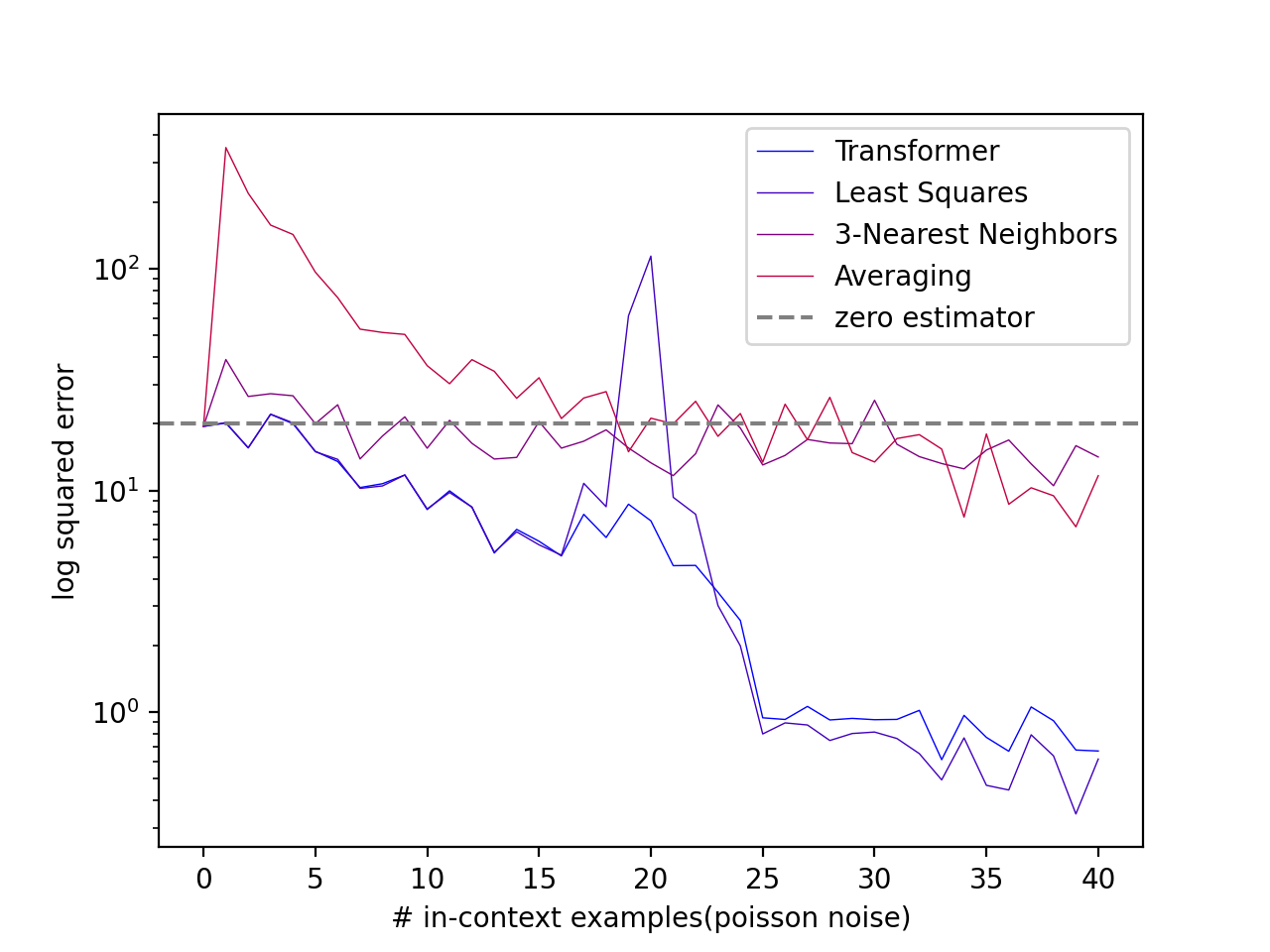} &
     \includegraphics[width=0.24\linewidth,valign=c]{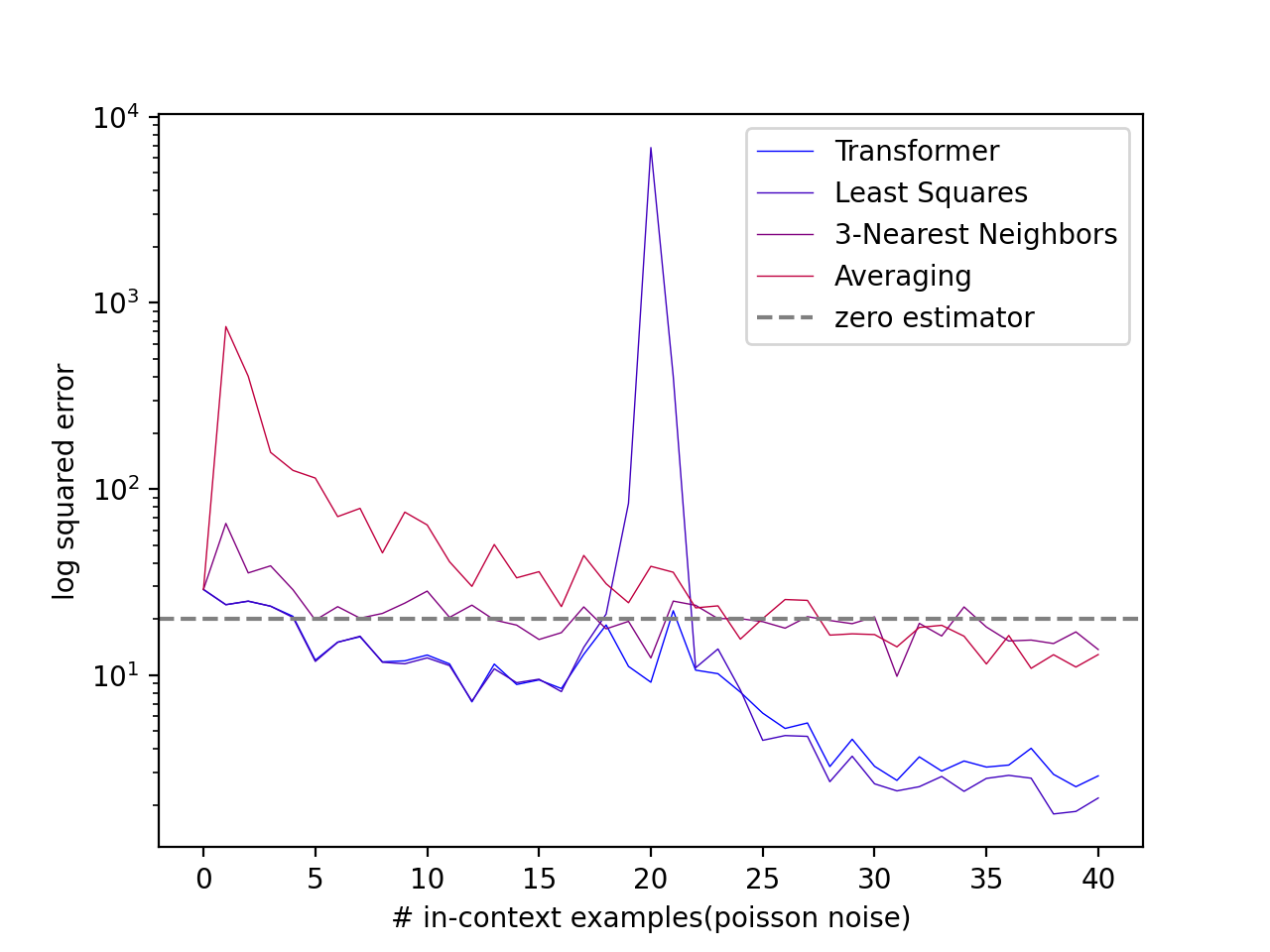} \\
     {\small Multiplicative} & \includegraphics[width=0.24\linewidth,valign=c]{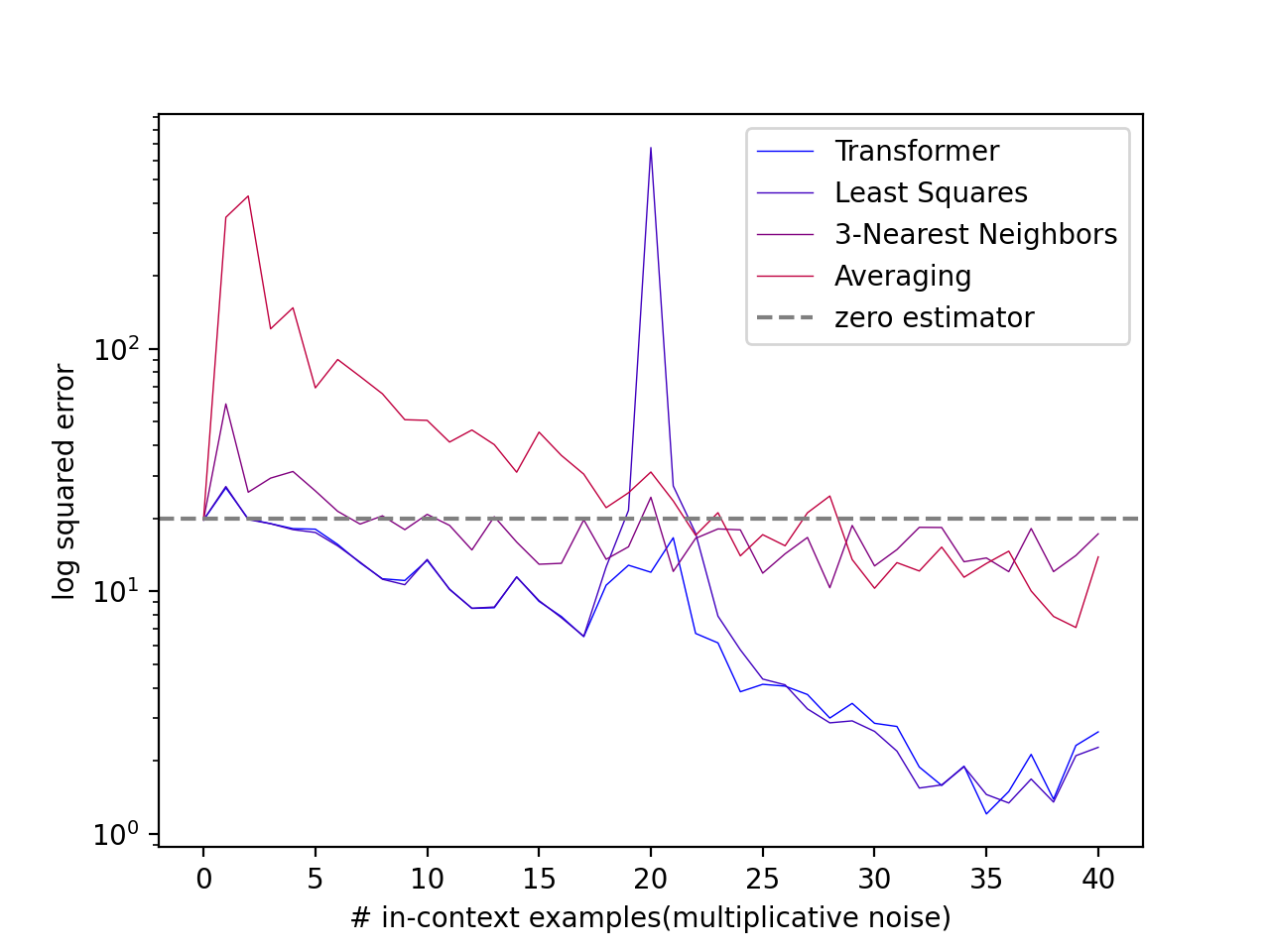} &
     \includegraphics[width=0.24\linewidth,valign=c]{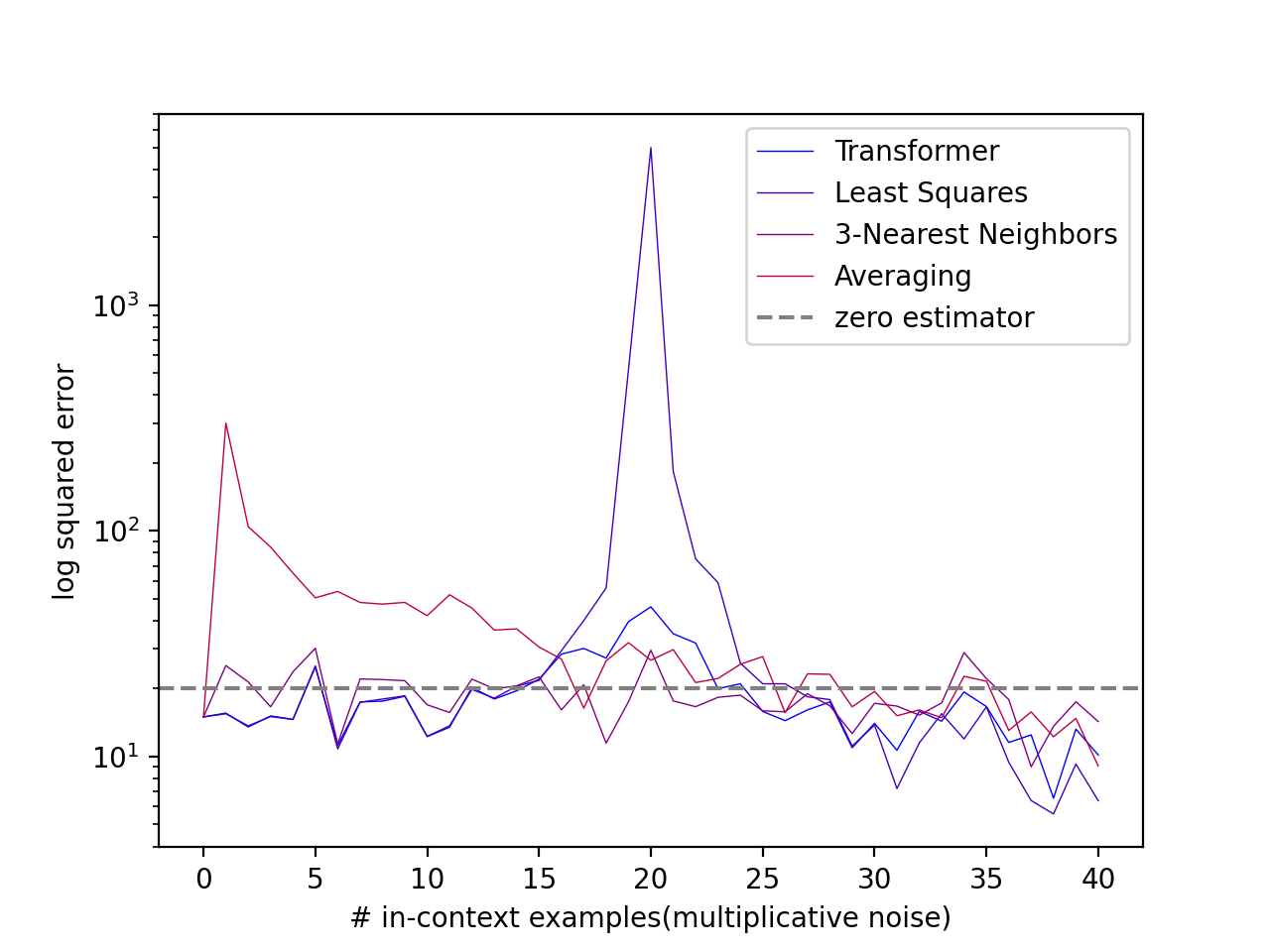} &
     \includegraphics[width=0.24\linewidth,valign=c]{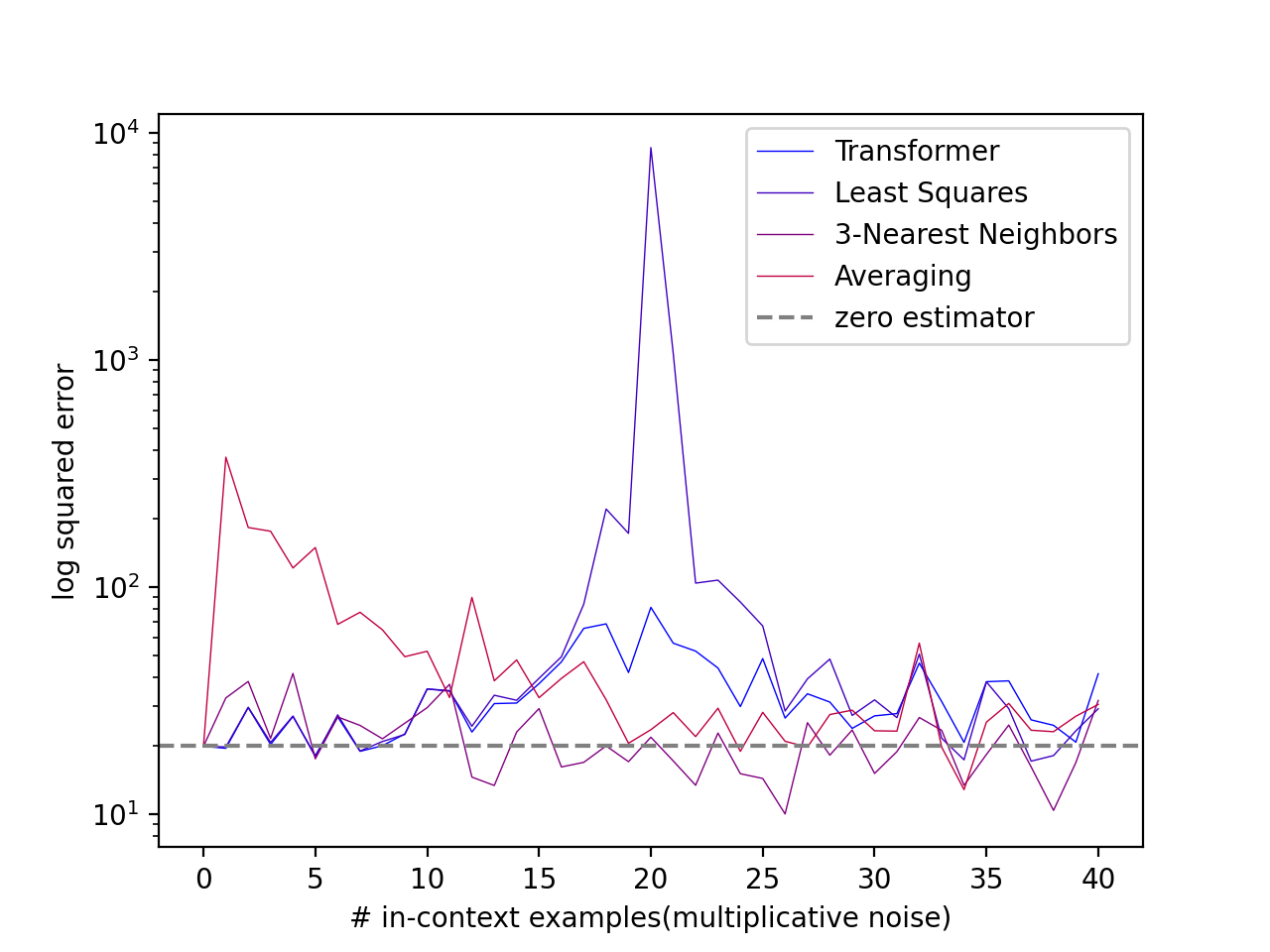} \\
     {\small S\&P} & 
     \includegraphics[width=0.24\linewidth,valign=c]{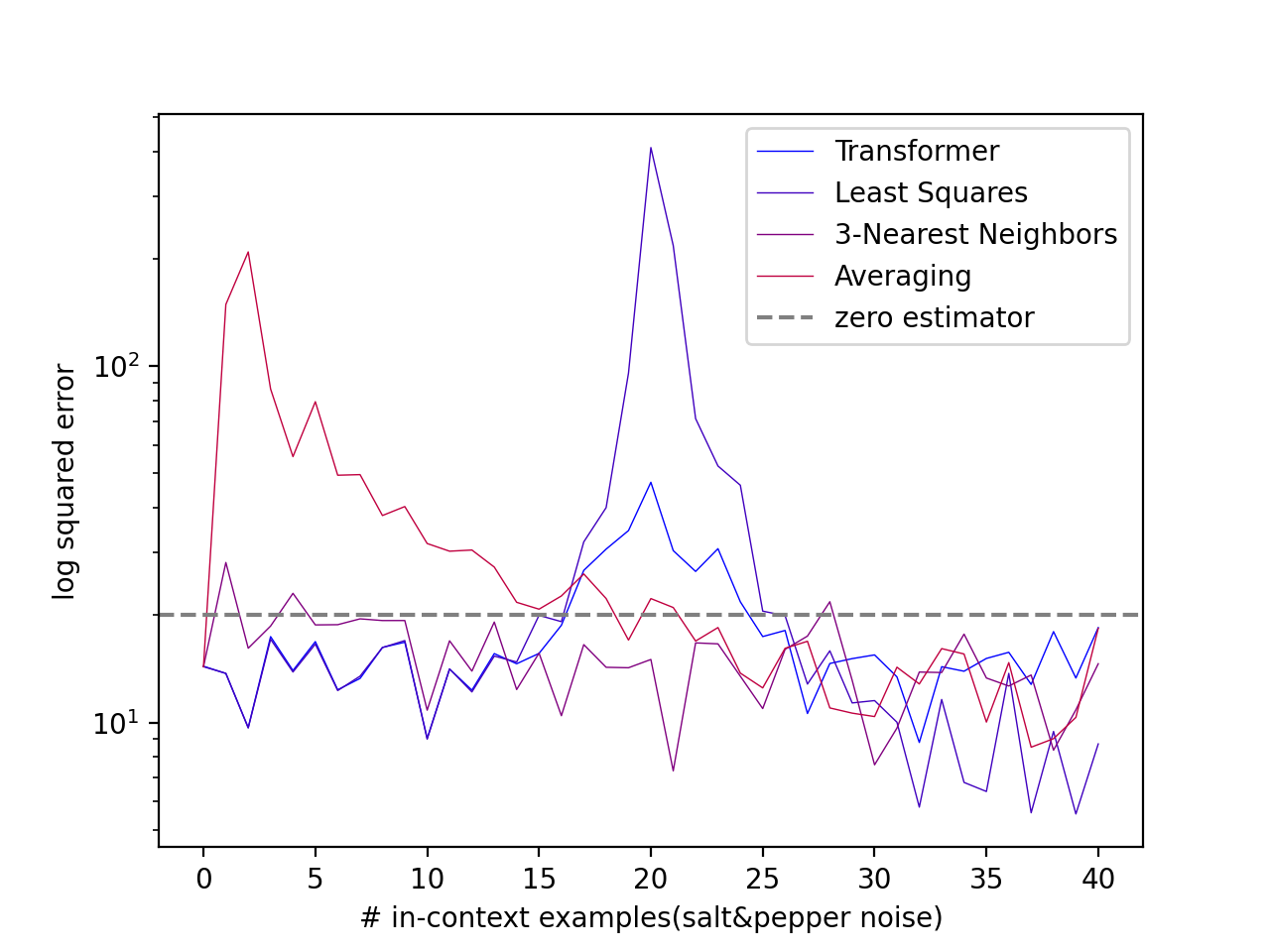} & \includegraphics[width=0.24\linewidth,valign=c]{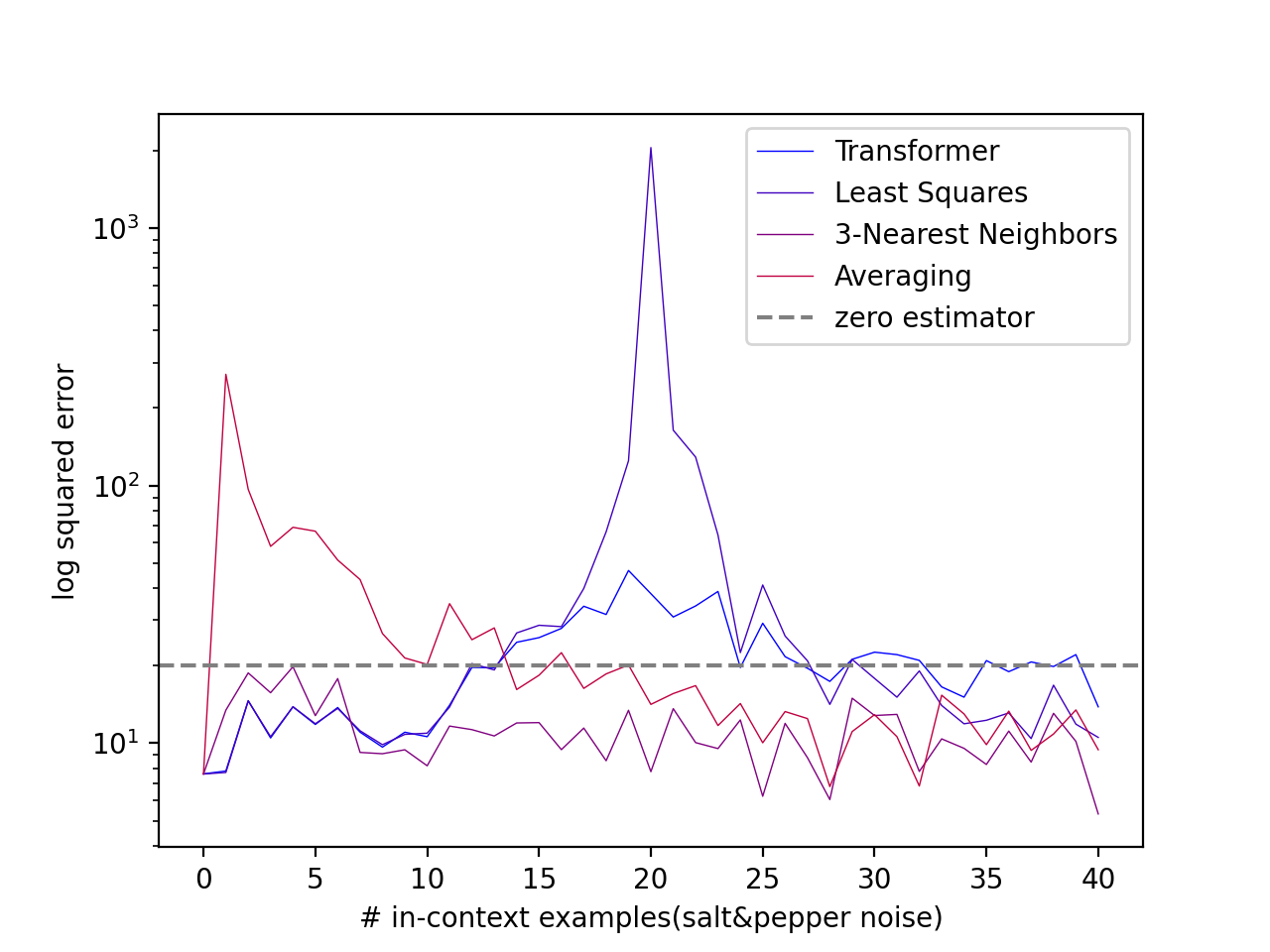} &
     \includegraphics[width=0.24\linewidth,valign=c]{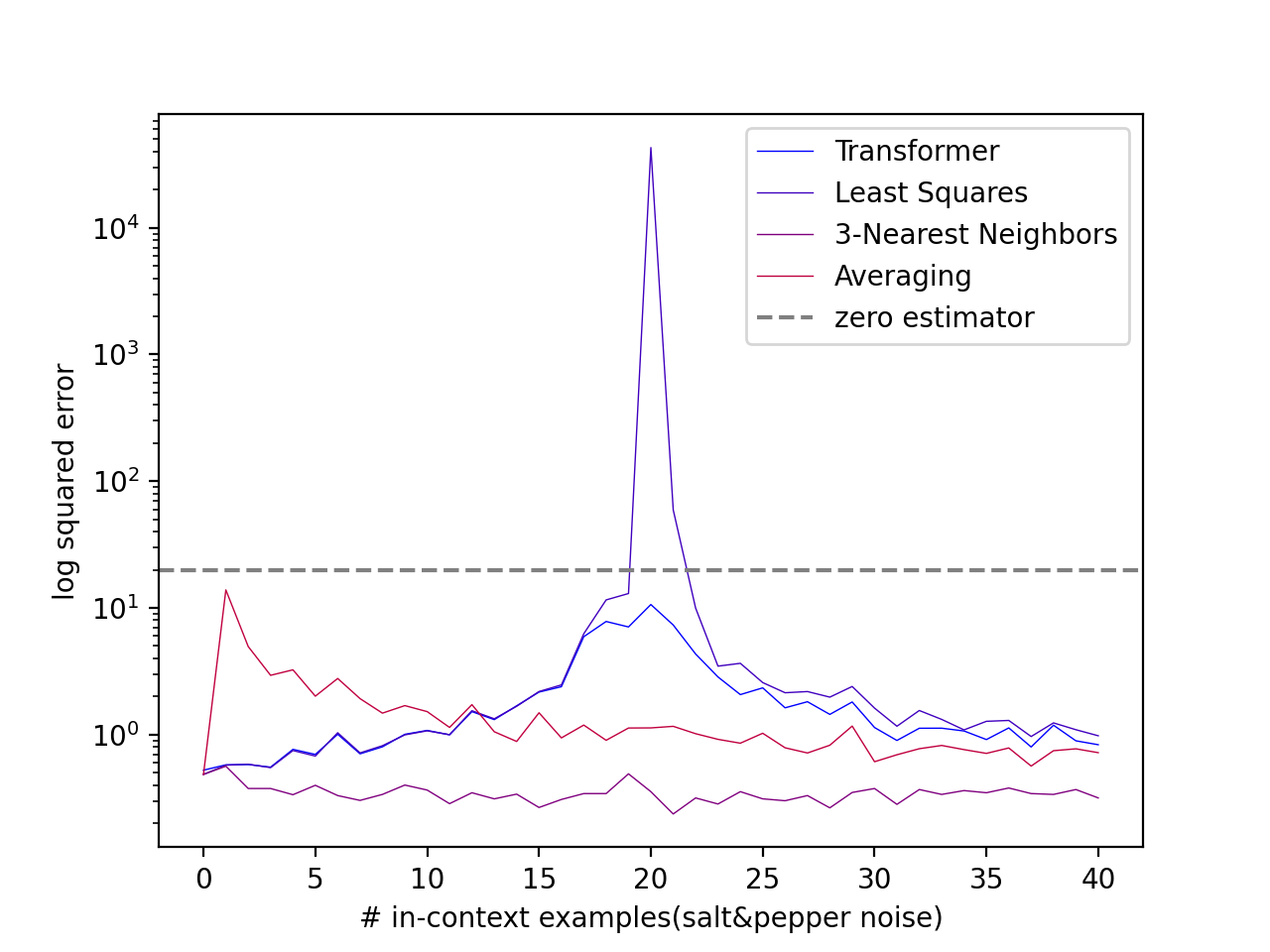} \\
    & $\sigma = 0.2$ &$\sigma = 0.5$ &$\sigma = 1.0$  
\end{tabular}
\caption{Robustness Comparison under Different Noise Types and magnitudes. Each figure represents a noise and each line represents a $\sigma_{test}$. {The X-axis represents the number of in-context examples.} 
}
\label{fig: 5_1}
\end{figure}

\section{Non-i.i.d. noises}
\label{sec Non iid}

{We further extend our experiments to the \textit{non-i.i.d.} noises scenario. Specifically, we replace the \textit{i.i.d.} label noises with some outlier demonstrations, which is a particular case of \textit{non-i.i.d.} noises, and analyze how the transformers respond to such \textit{non-i.i.d} noise in the demonstrations.}

\noindent\textbf{Experiment set-up.} In this experiment, we still focus on the class of noisy linear functions $F=\{f_w | f_w(x) = w ^Tx +\mathbb{1}_{outlier} *\epsilon, w \in R^d\}$ in $d=20$ dimensions, where noises are only injected into demonstrations of \textbf{outliers}. Similar to the construction of the noisy demonstrations, we sample $x_1,\cdots, x_k, x_{query}$ from the isotropic Gaussian distribution $N(0, I_d )$. Then, we randomly select $m$ outliers $\mathbb{S}=\{i_1,\cdots,i_m\}$, and generate \textit{i.i.d.} Poisson noise $\{\epsilon_{i_1},\cdots,\epsilon_{i_m}\}$ with noise magnitude $\sigma$. Subsequently, each $y_i = w^Tx_i +\mathbb{1}_{i\in S} *\epsilon_i$ is computed. Finally, the prompt $P$ is constructed as $P = (x_1, y_1, x_2, y_2, \cdots , x_k , y_k , x_{query})$.

\noindent\textbf{Robustness analysis.}
We present the results with small multiples in Figure \ref{fig: outliers}. Overall, the transformer model shows superior performance than other baselines, which is similar to the results of \textit{i.i.d.} noises presented above and shows its consistent robustness against such noises, {which we elaborate on below.}

\begin{figure*}[h]
    \centering
    \includegraphics[width=0.98\linewidth]{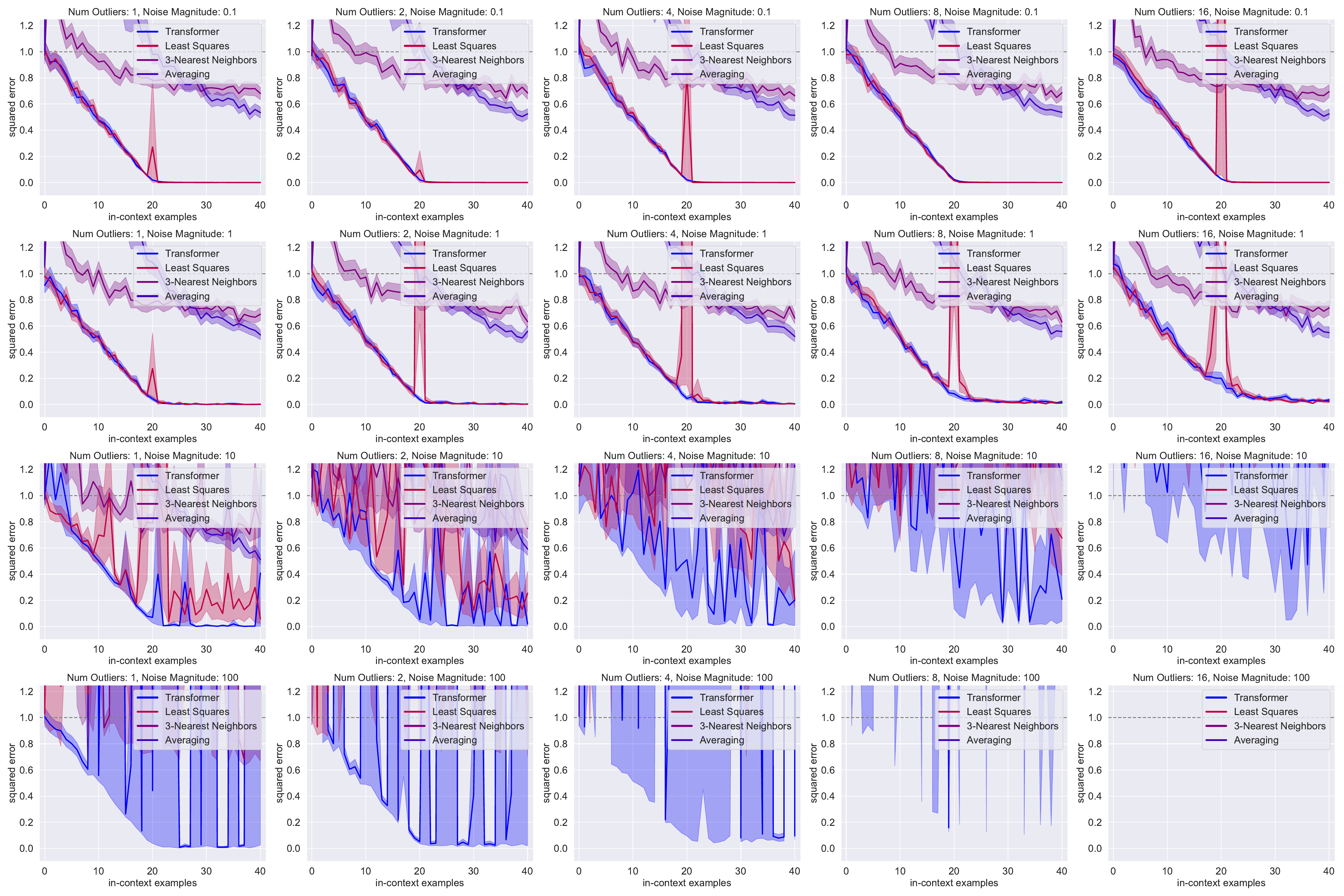}
    \caption{In-context learning on prompts with outliers. We evaluate the trained model on prompts with outliers in two cases. (1) On the rows, we explored the effect of the number of outliers (\{1,2,4,8,16\}), and (2) on the columns, we explored the effect of the magnitude of outliers (\{0.1,1,10,100\}). (Error averaged over 1280 prompts. 90\% confidence intervals over 1000 bootstrap trials.) }
    \vspace{-10pt}
    \label{fig: outliers}
\end{figure*}

\noindent\textbf{Outliers of different magnitudes and equal quantity.} 
 {Analyzing different columns in Figure~\ref{fig: outliers} where the noise of varying magnitudes was introduced to a specified number of prompt outputs, we can see that the trained model exhibits greater robustness and stability compared to baseline models, particularly as the noise magnitude escalates.} Notably, the lower bound of the 0.9 confidence interval closely aligns with the actual data when both the noise magnitude and the number of outliers are minimal, suggesting that the model maintains high prediction accuracy 
 under these conditions.
 
\noindent\textbf{Different quantities of outliers with uniform magnitude.} 
On the other hand, from different rows in Figure~\ref{fig: outliers} where the sub-figures involve different numbers of outliers and a consistent magnitude within the prompts, we can see an increase in the trained model's robustness and stability as the number of outliers rises.

\section{Different function complexities}
\label{sec function}

In this part, to evaluate the transformer's ICL ability in terms of capturing more complex patterns within input-output mappings, which may be closer to real-world data,
we broaden our exploration beyond linear relationships by exploring the task of \textbf{polynomial regression} ranging from degrees of 2 to 5. We compare the transformer's robustness with both the least-squares method and a 3-NN model. 

\noindent\textbf{Experiment Set-up.} 
We continue to adopt the GPT-2 model architecture. To construct a noisy training set, instead of vanilla demonstrations $\{x_i, f_w(x_i)\}$, we introduce Gaussian noise to the labels as $\{x_i, f_w^{(n)}(x_i) + \epsilon_i\}$, where
\begin{equation}
    f_w^{(n)}(x_i) = \sum_{i=1}^n w_i \cdot x_i^n,
\end{equation}
and $a_i$ is sampled from $N(0, 1)$.
Here, $\epsilon_i \sim \mathcal{N}(0, \sigma^2)$, and we consider $\sigma\in  \{0.1,0.2,0.5\}$. 

\noindent\textbf{Baselines.}
We introduce two baseline methods for evaluating the transformer's ability to model polynomial objective functions: 

\emph{(1) Least Squares Model (LSM):} The Ordinary Least Squares (OLS) regression constructs polynomial features up to the specified degree n for each input sample and computes the coefficients $\mathbf{w}$ using the least squares optimization $\mathbf{w} = (\mathbf{X}^\intercal \mathbf{X})^{-1} \mathbf{X}^\intercal \mathbf{y}$ where $\mathbf{X}$ is the design matrix consisting of polynomial features, $\mathbf{y}$ is the vector of labels, and $\mathbf{w}$ are the coefficients. 

\emph{(2) k-Nearest Neighbors Model (k-NN):} We extend the k-Nearest Neighbors algorithm applied in the experiments above with polynomial features. For each test point $x_{\text{test}}$, the predicted output $\hat{y}$ is computed by averaging the labels of the 3 nearest neighbors.

\begin{figure}[h]
    \centering
        \label{poly}
        \begin{tabular}{cccc}
        \includegraphics[width=0.23\linewidth]{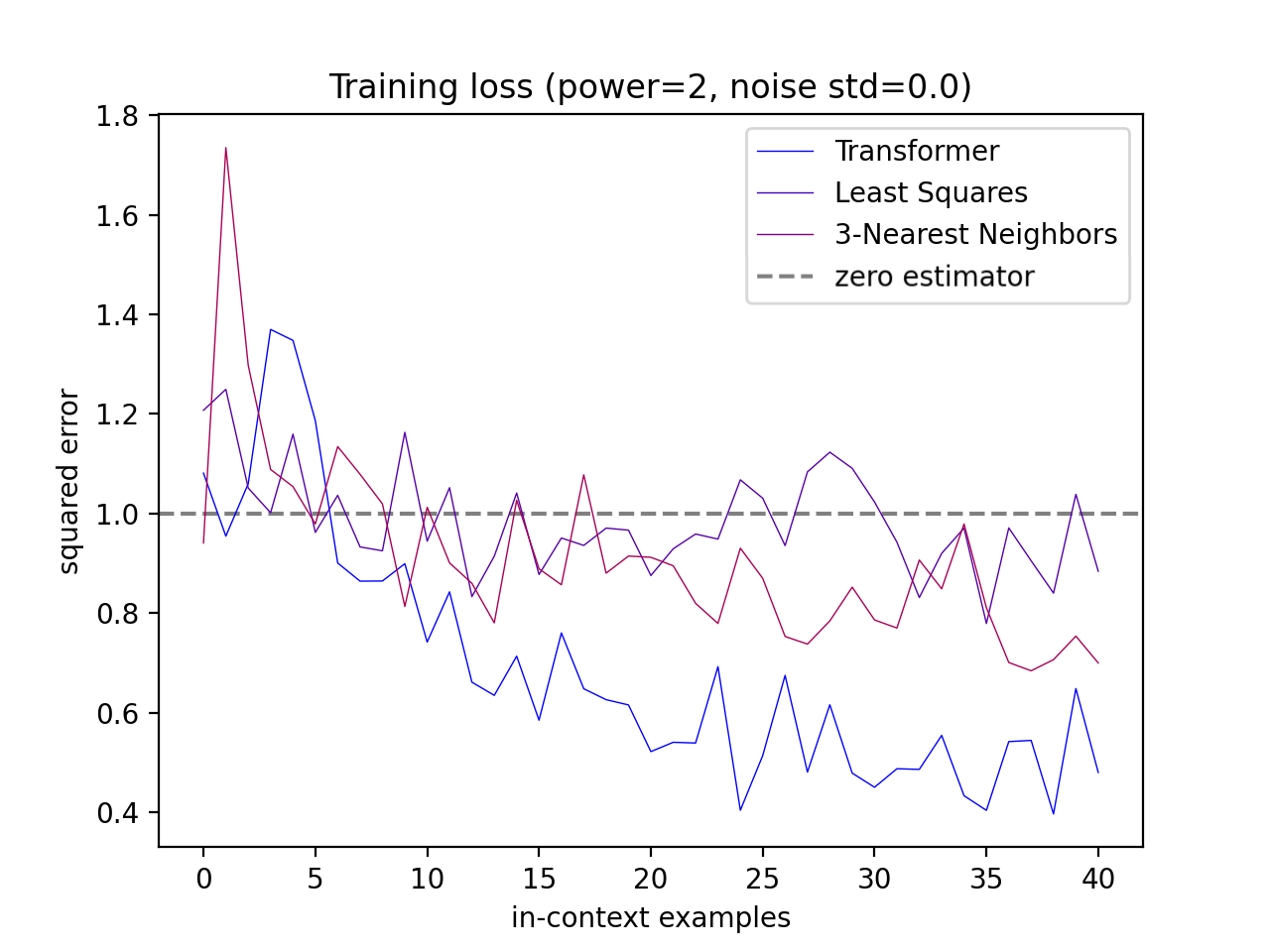}
         & \includegraphics[width=0.23\linewidth]{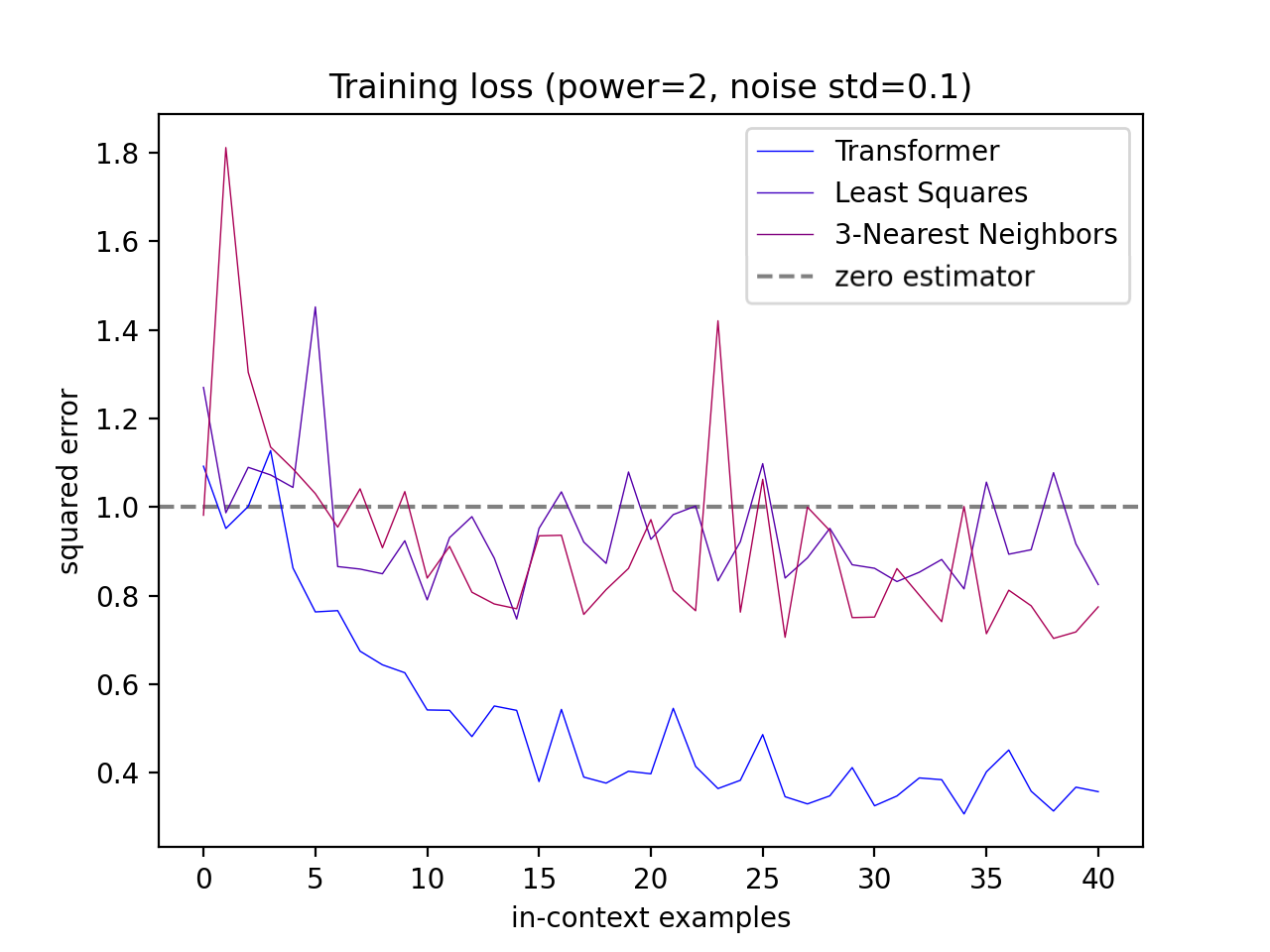}
         & \includegraphics[width=0.23\linewidth]{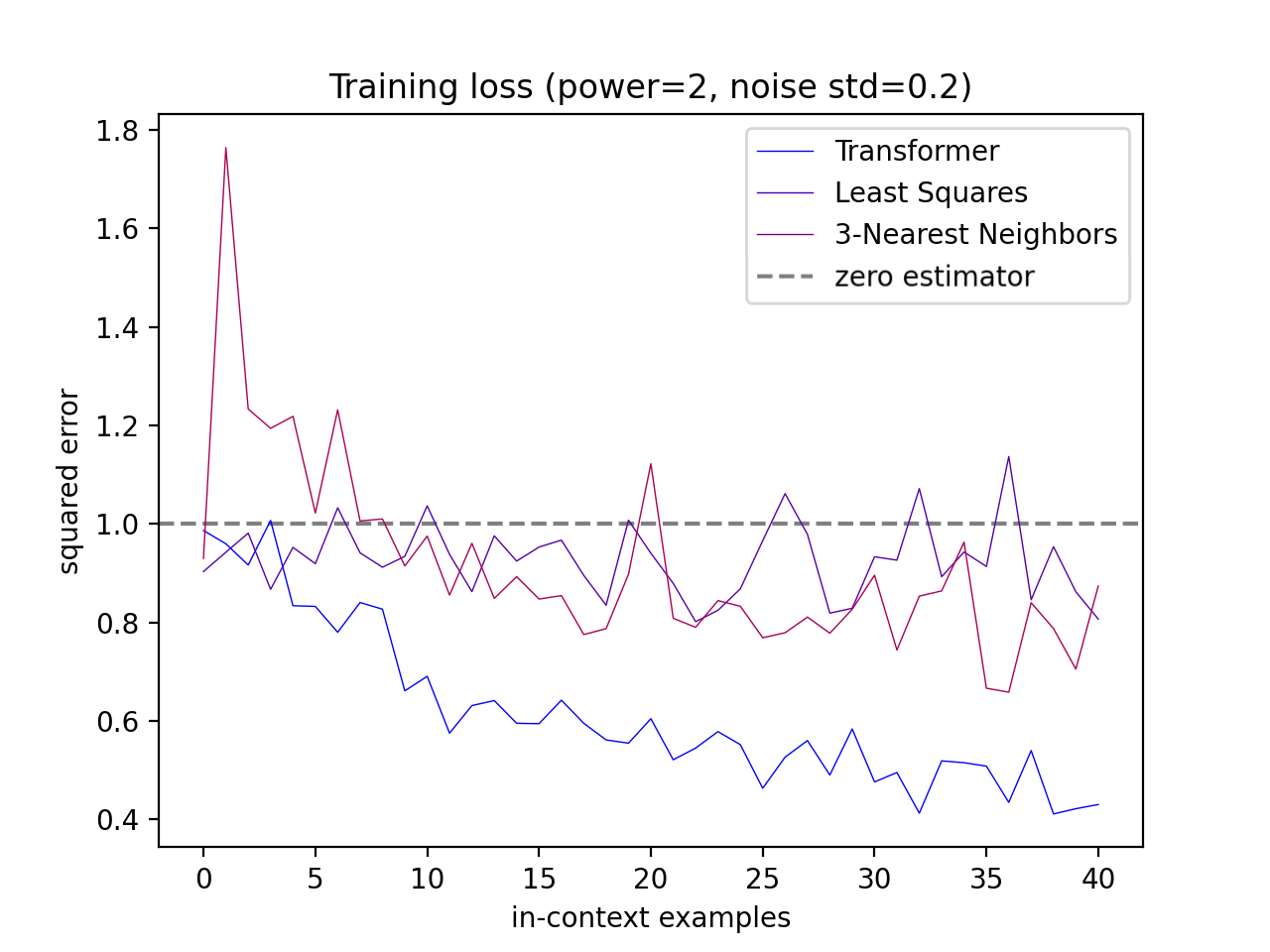}
         & \includegraphics[width=0.23\linewidth]{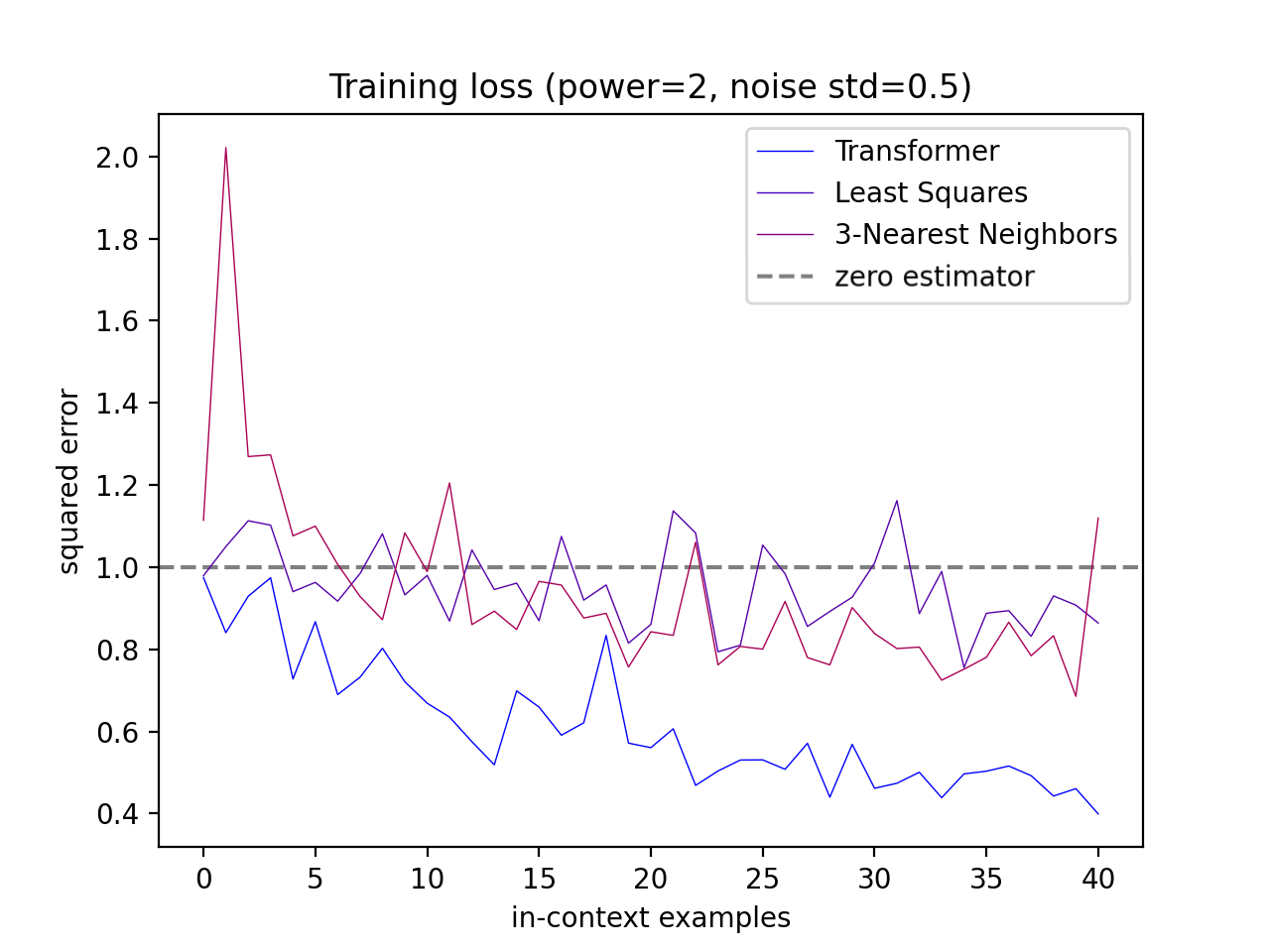}
        \\
        \includegraphics[width=0.23\linewidth]{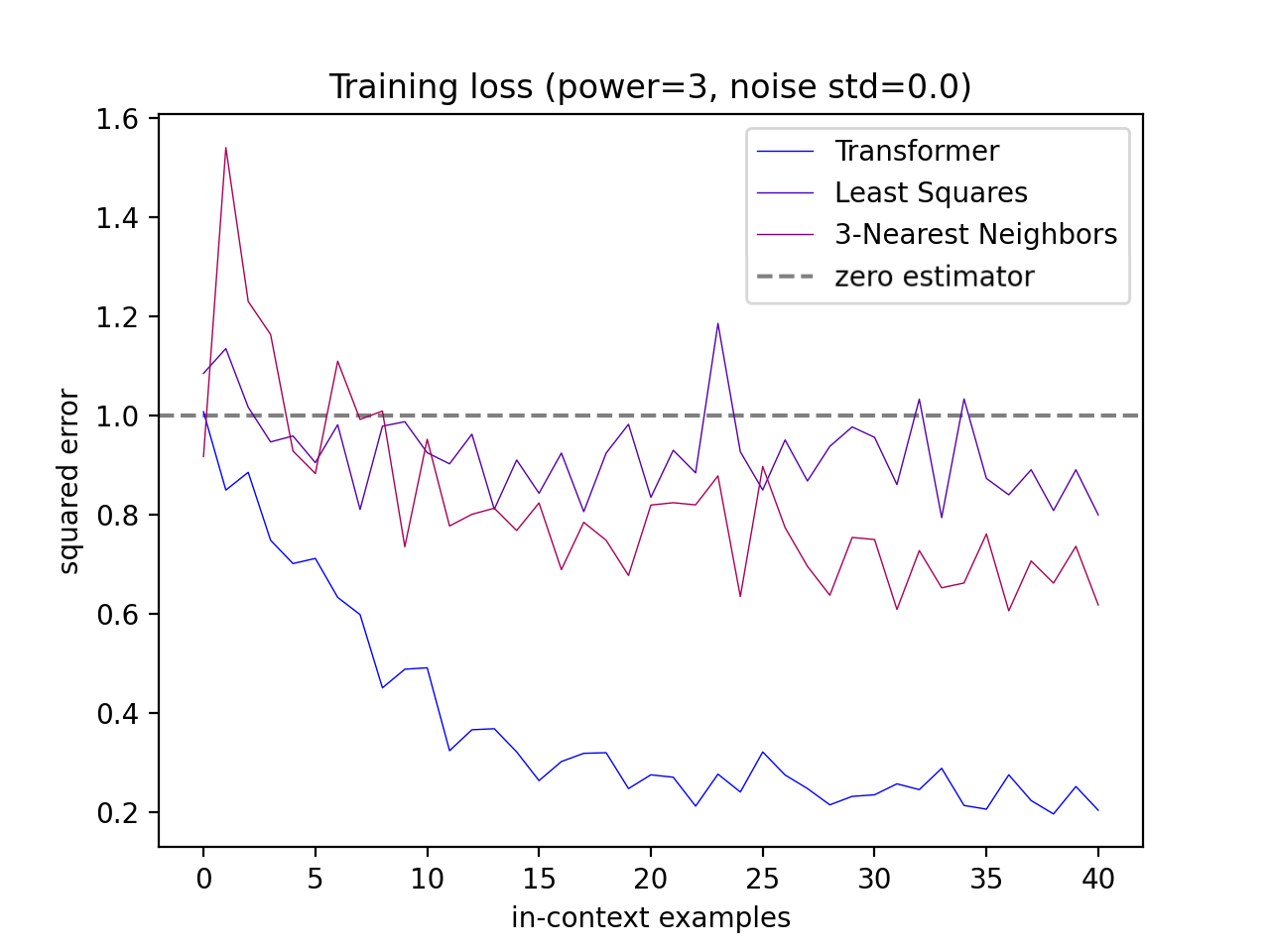}
         & \includegraphics[width=0.23\linewidth]{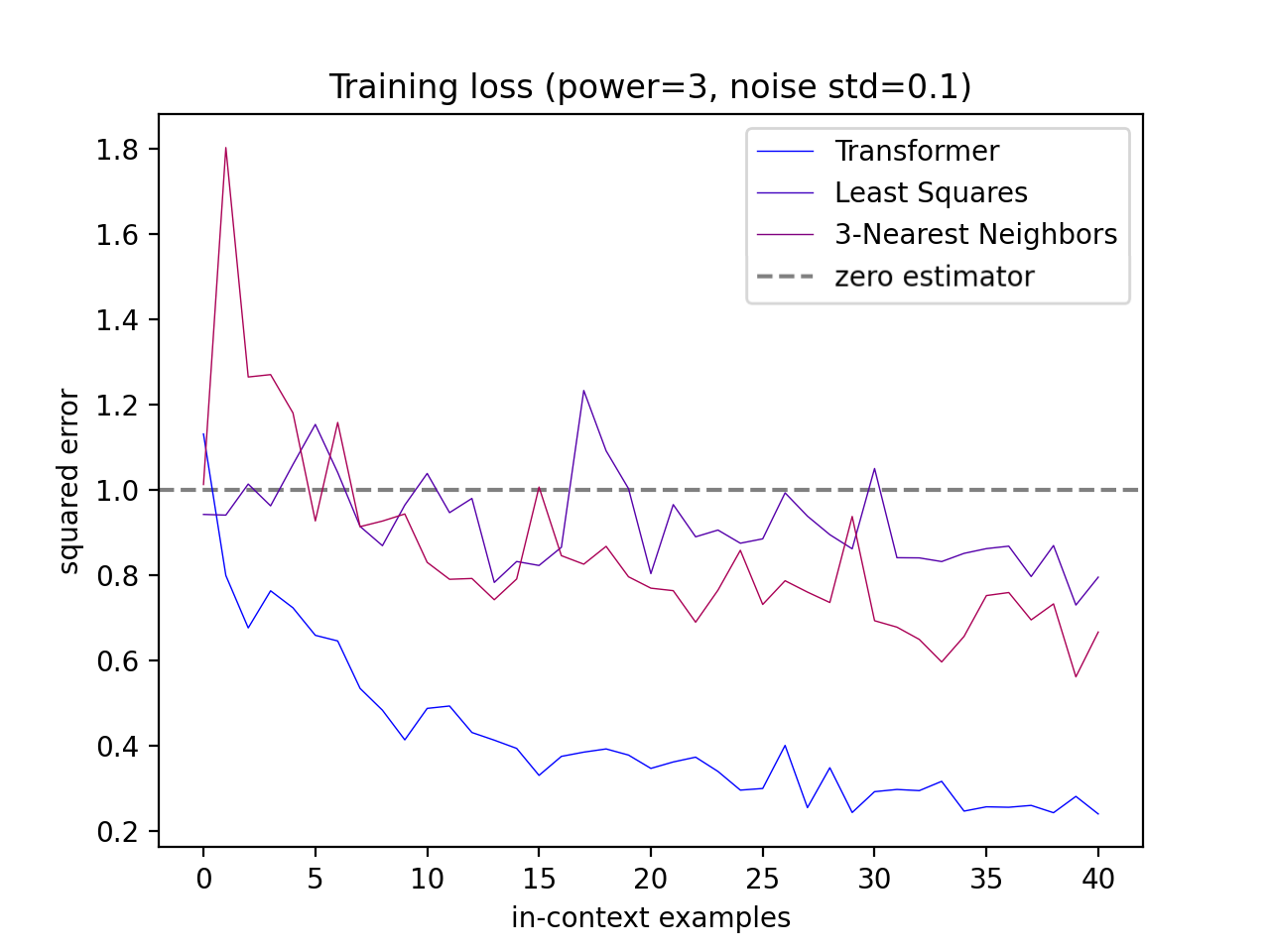}
         &
        \includegraphics[width=0.23\linewidth]{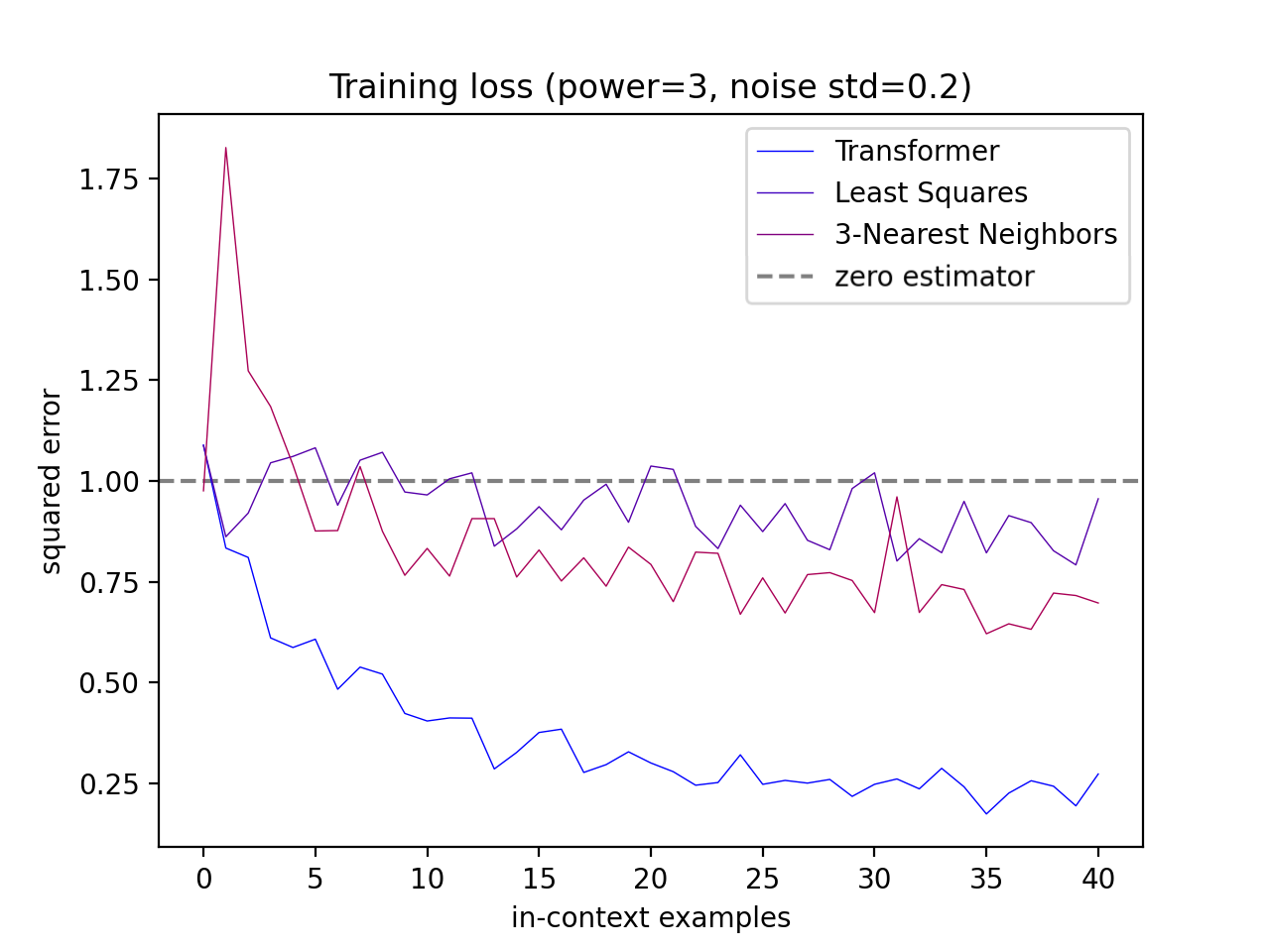}
        & \includegraphics[width=0.23\linewidth]{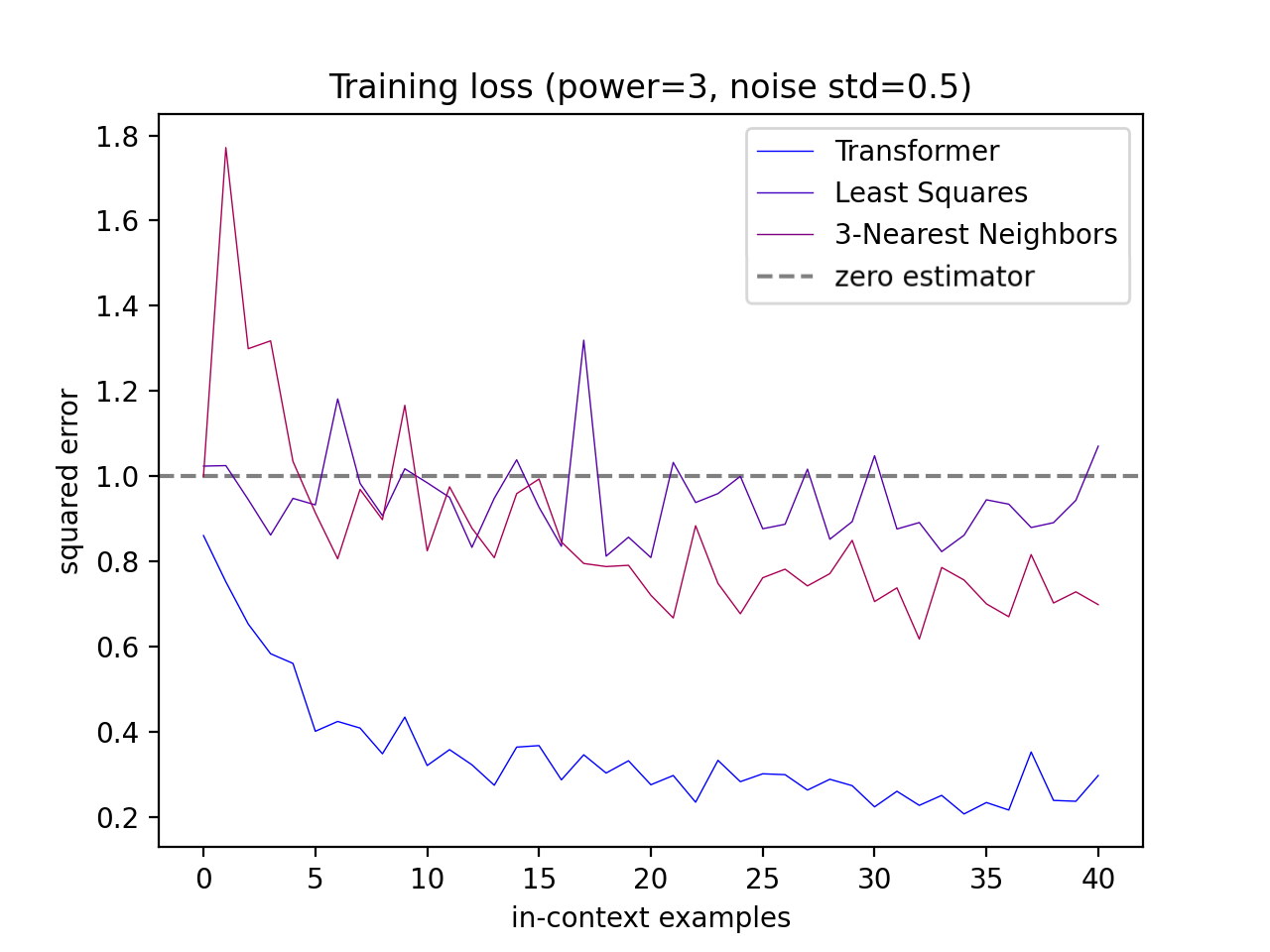}
        \\
        \includegraphics[width=0.23\linewidth]{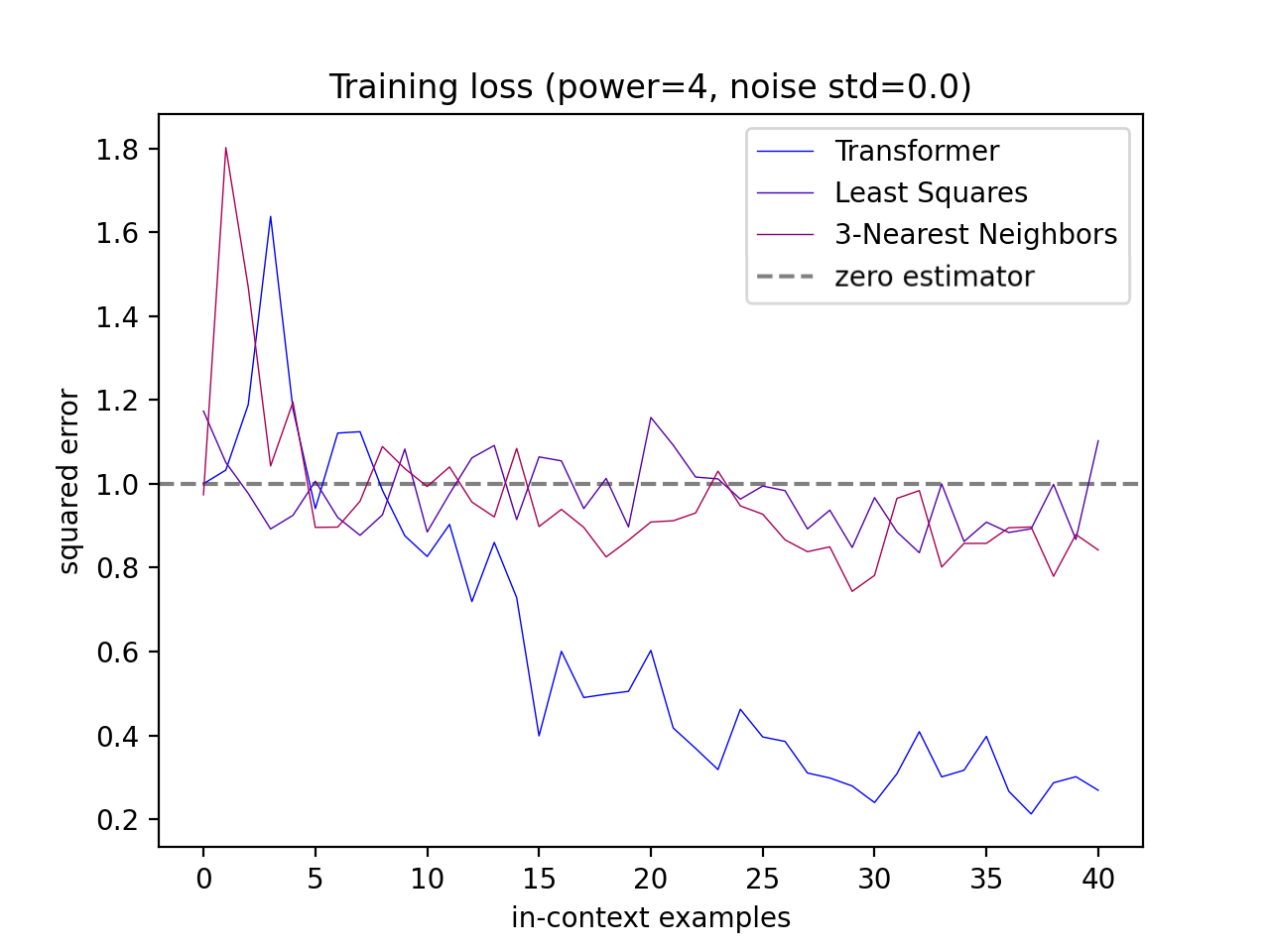}
         & \includegraphics[width=0.23\linewidth]{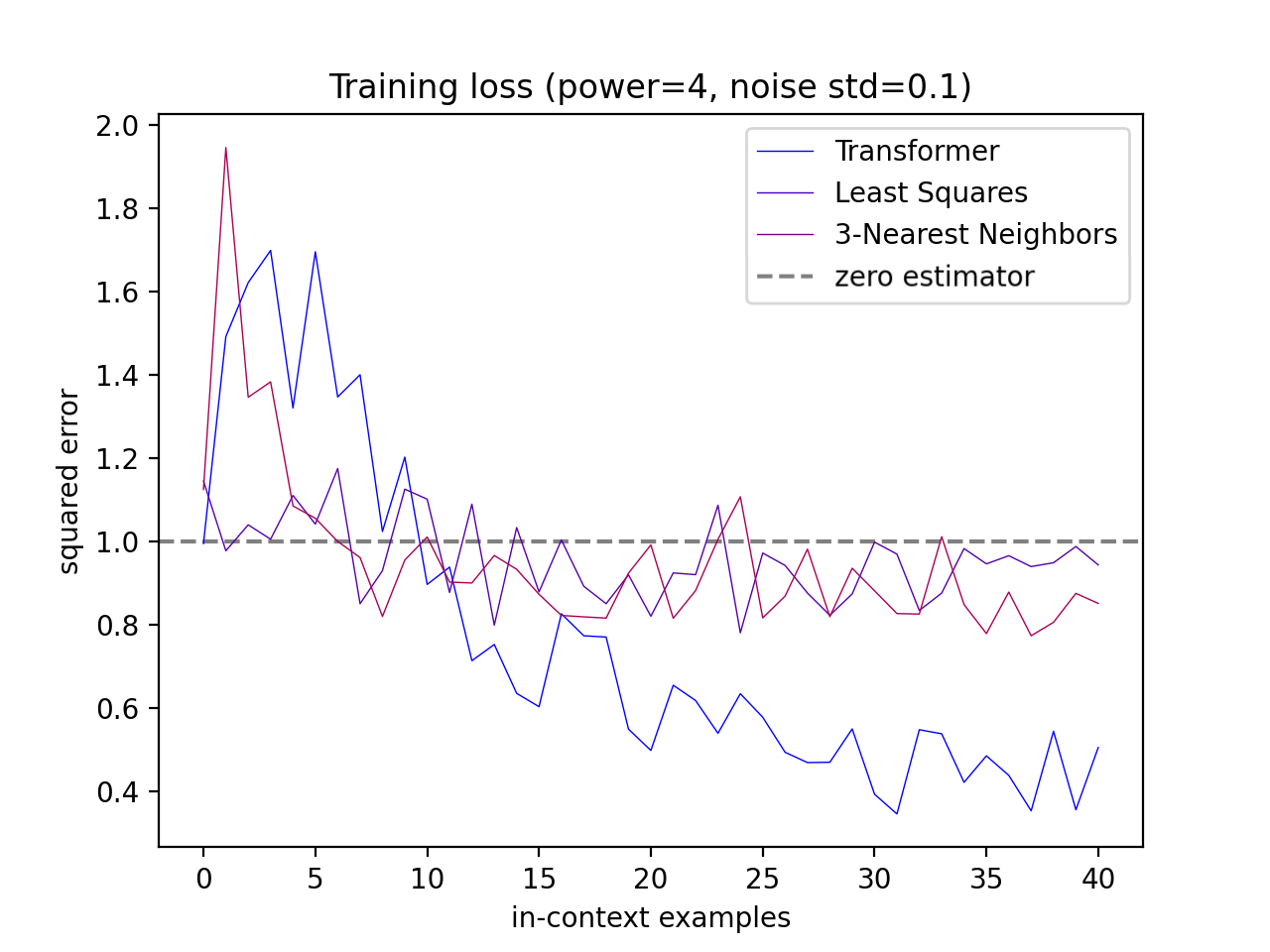}
         &
        \includegraphics[width=0.23\linewidth]{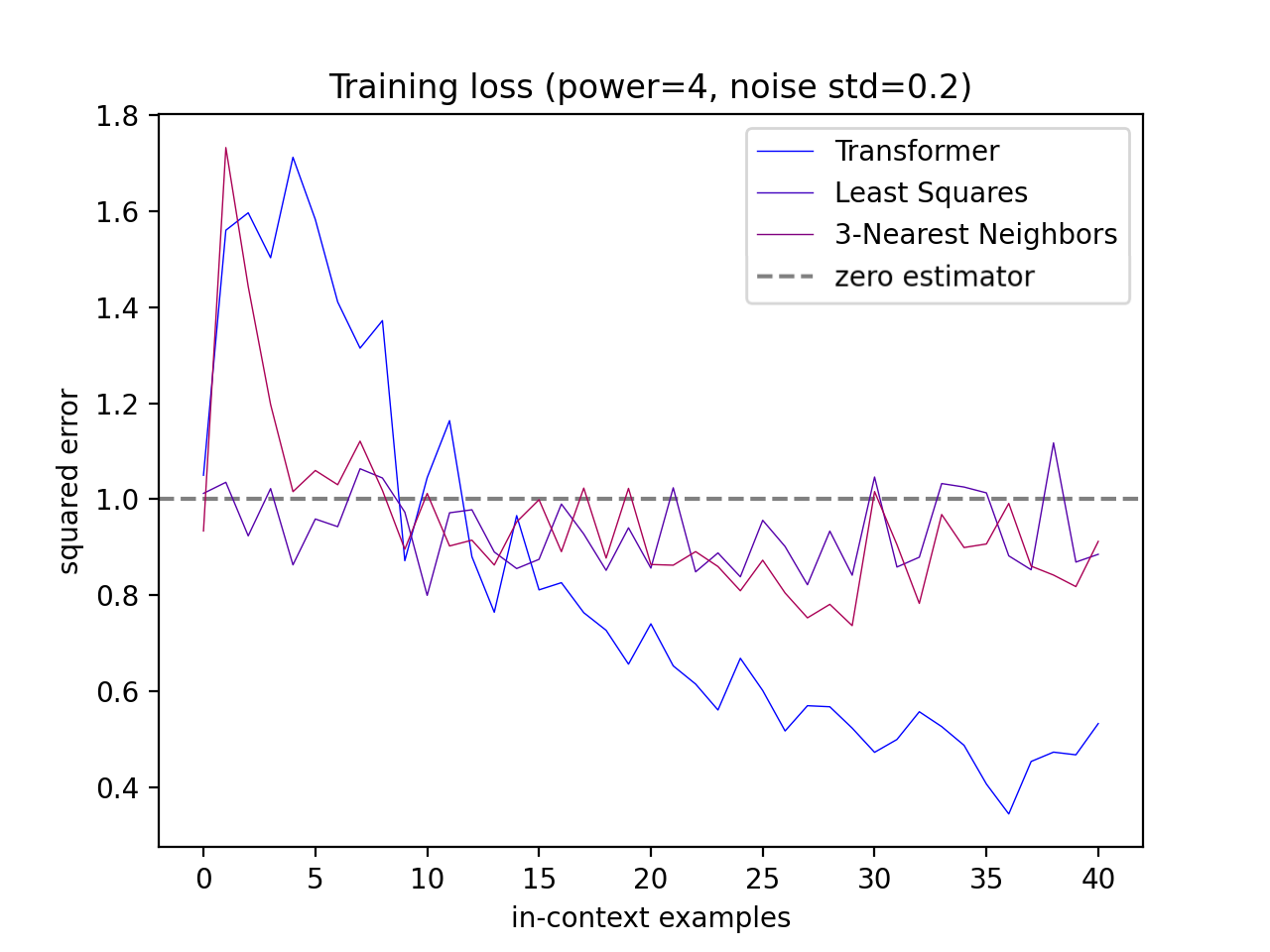}
        & \includegraphics[width=0.23\linewidth]{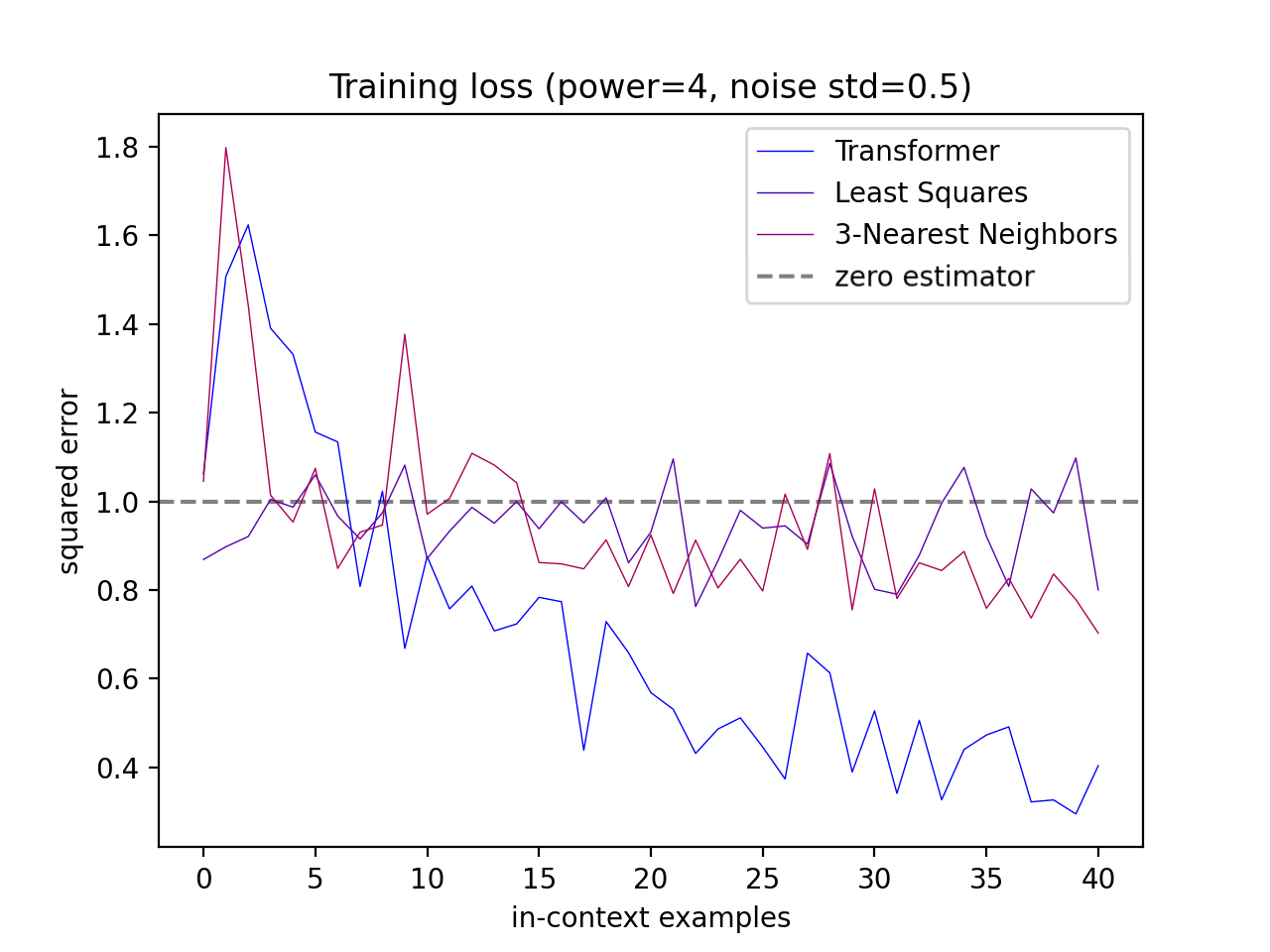}
        \\
        \includegraphics[width=0.23\linewidth]{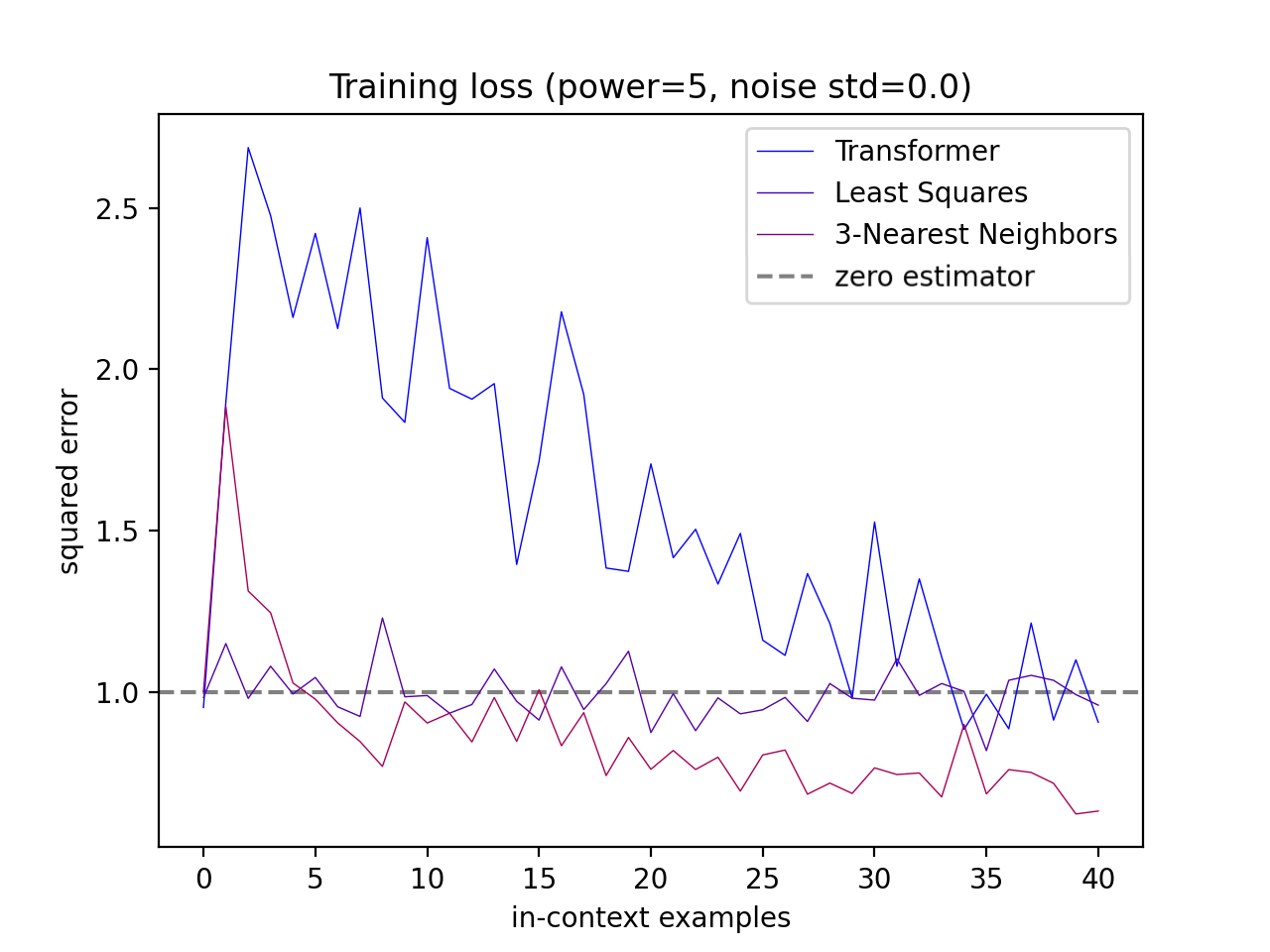}
         & \includegraphics[width=0.23\linewidth]{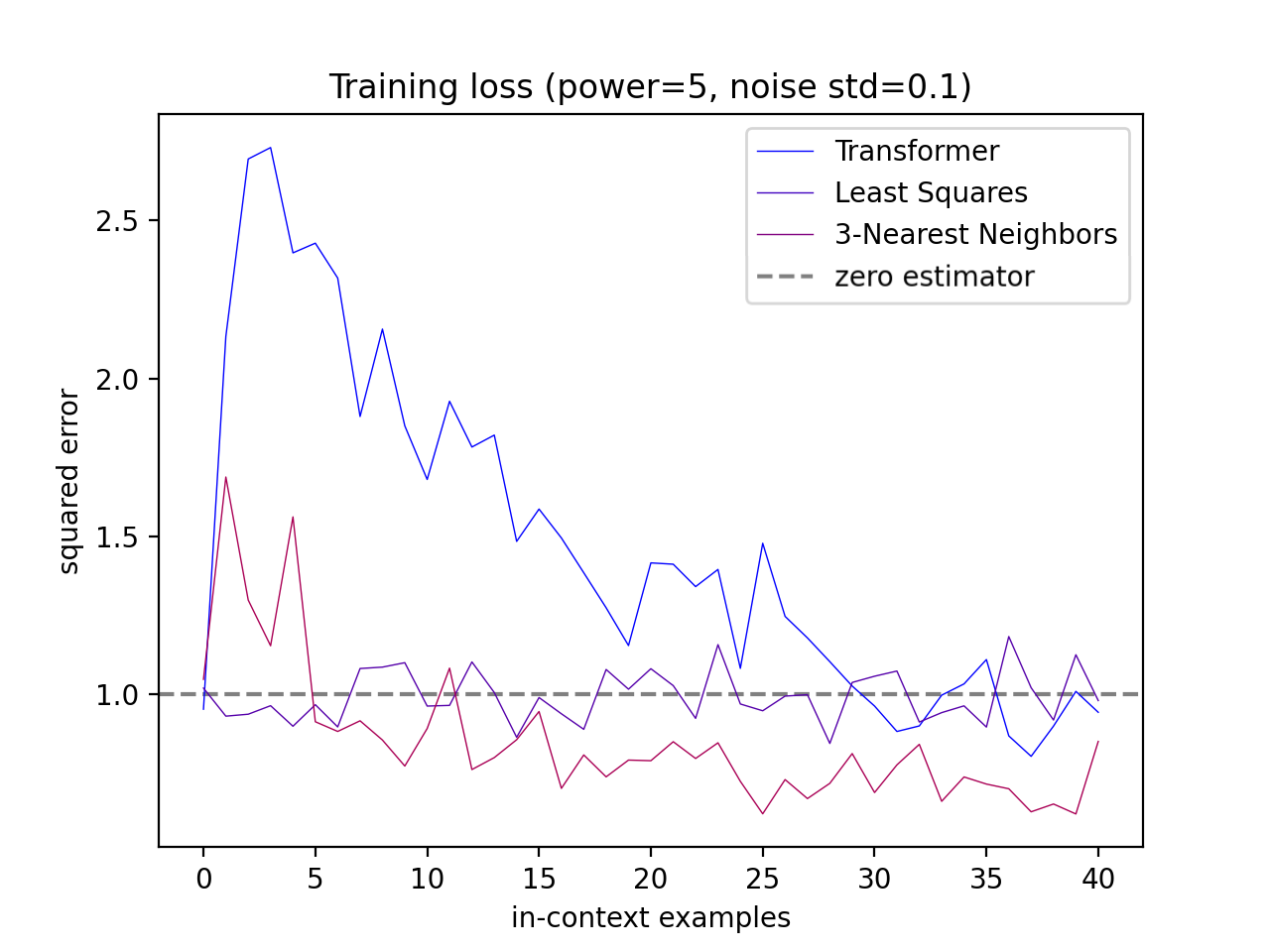}
         &
        \includegraphics[width=0.23\linewidth]{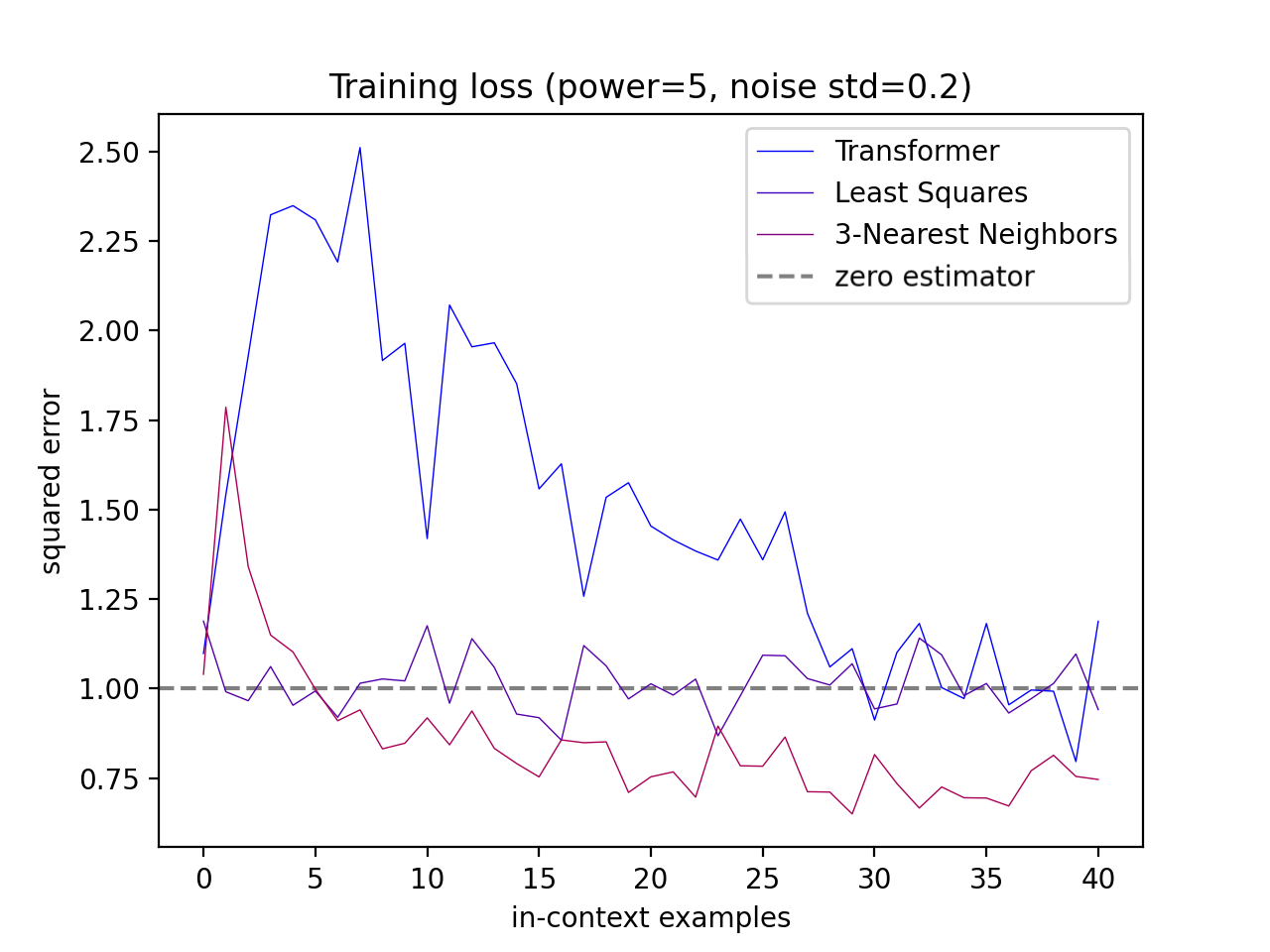}
        & \includegraphics[width=0.23\linewidth]{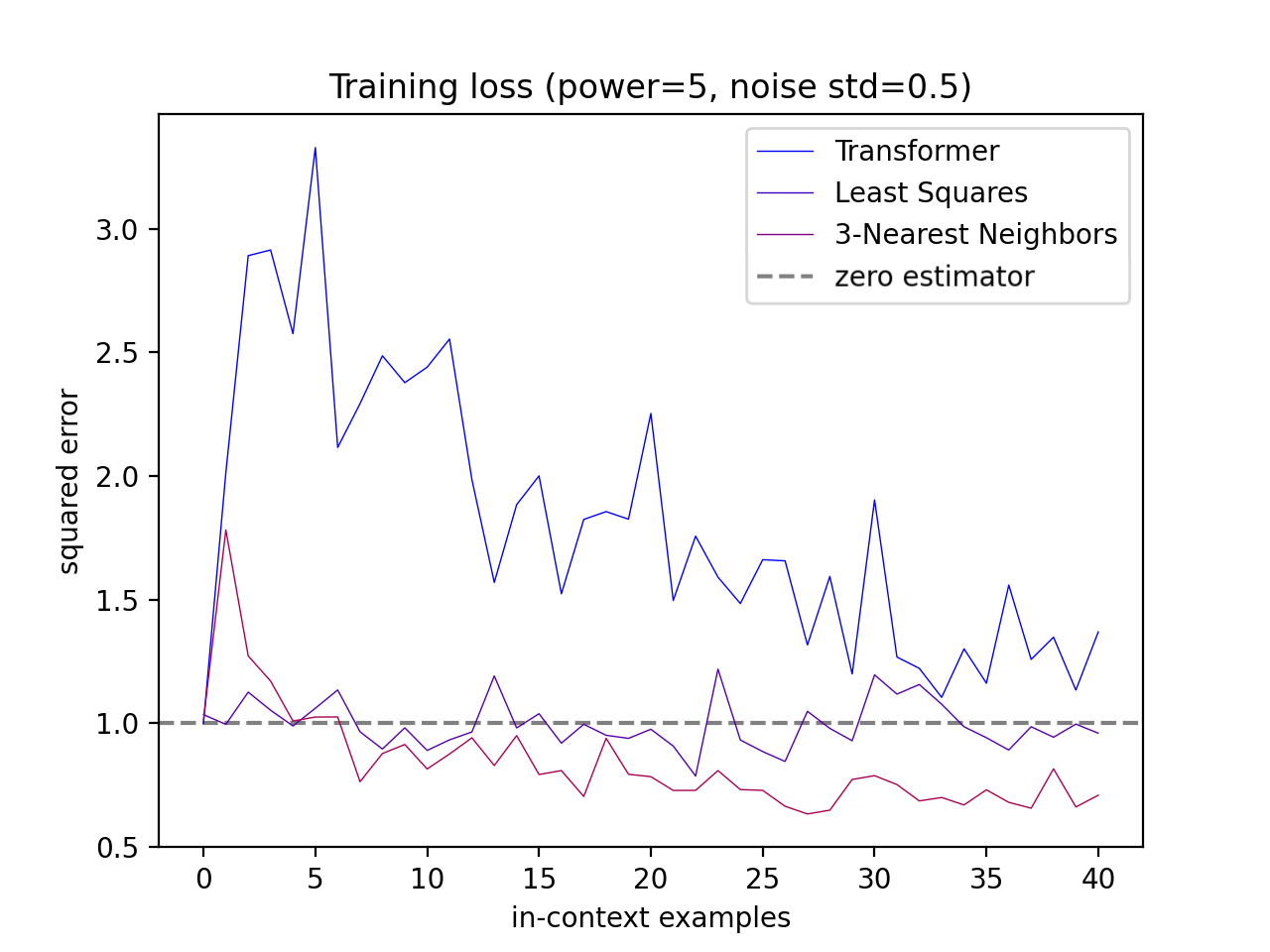}
        \\
        $\sigma_{\text{test}}=0.0$ & $\sigma_{\text{test}}=0.1$ & $\sigma_{\text{test}}=0.2$ & $\sigma_{\text{test}}=0.5$
        \\
    \end{tabular}
    \caption{\textbf{noisy ICL performance} comparison for models trained with different polynomial degrees. {Each figure represents an inference noise level $\sigma_{test}$, and each line represents a polynomial degree. The X-axis represents the number of in-context examples.}}
\end{figure}

\noindent\textbf{Robustness analysis.}
We evaluate the Transformer model's performance on polynomial functions with degrees $g$ from 2 to 5 and plot the results in Figure~\ref{poly}. We found that while the Transformer consistently outperformed baseline models for $g=2$ and  $g=3$, its performance varied as polynomial complexity increased. For objectives of $g=4$, the Transformer initially lagged but eventually surpassed baselines significantly. However, for objectives of $g=5$, its efficacy declined compared to the baselines. 

This performance disparity can be attributed to several factors. Firstly, the heightened polynomial degree engenders an escalated functional complexity, posing formidable challenges to the Transformer's capacity to model intricate higher-order nonlinear relationships. Secondly, the dearth of adequately diverse examples may exacerbate performance degradation, impeding the Transformer's ability to accurately capture and characterize complex functions due to insufficient training data. This limitation underscores the Transformer's susceptibility to "in-context" learning deficits when confronted with inadequate exemplars. Lastly, the phenomenon of overfitting may further compound performance disparities, particularly evident in the context of degree 5 objectives.

Notwithstanding these challenges, our comprehensive analysis underscores the Transformer model's commendable performance in addressing polynomial objective functions encountered in routine real-world scenarios. Its demonstrated prowess in generalization and robustness, particularly evident in outperforming baseline methods for lower-degree polynomials, underscores its potential efficacy in diverse applications and its capacity to navigate intricate nonlinear modeling tasks.

\section{Different input dimensions}
\label{sub set 4 dim}

In this section, we change the input vector dimension $d$ to see how the robustness of the model varies with it. 

\noindent\textbf{Experiment Set-up.}
In this experiment, we still focus on the class of linear functions $F=\{f_w | f_w(x) = w ^Tx + \epsilon, w \in R^d\}$, where $d$ varies in $\{3,5,10\}$, and $\epsilon \sim \mathcal{N}(0,\sigma^2)$. Specifically, we train the transformer model with different input dimensions, then we add Gaussian noise $\sigma \in \{0, 0.2, 0.4, 0.6, 0.8, 1\}$ during evaluation to study the influence of different dimensions on the robustness of Transformers.
\begin{figure}[h]
    \centering
    \begin{tabular}{ccc}
    
        \includegraphics[width=0.28\textwidth]{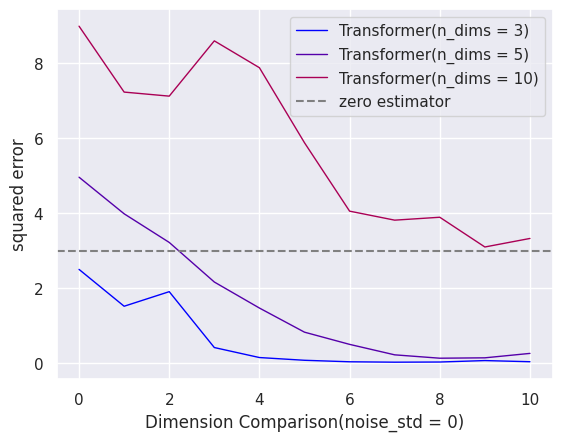}
         & \includegraphics[width=0.28\textwidth]{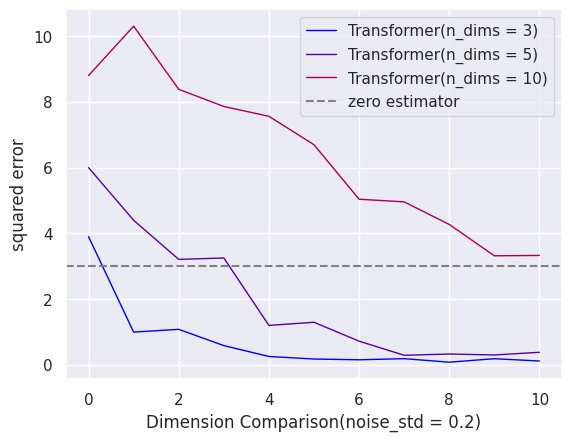}
         &
        \includegraphics[width=0.28\textwidth]{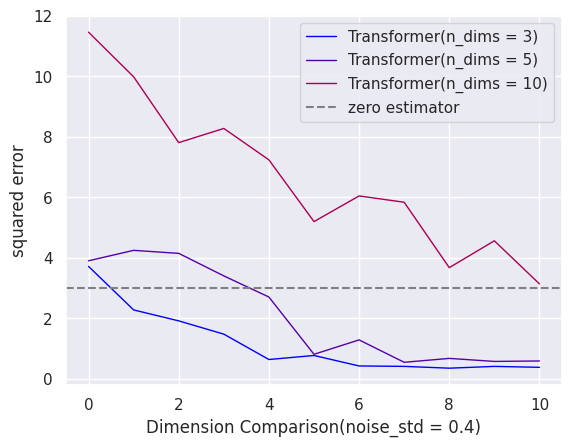}
        \\
        (a) $\sigma_{\text{test}}=0.0$ & (b) $\sigma_{\text{test}}=0.2$ &
        (c) $\sigma_{\text{test}}=0.4$ 
        \\
         \includegraphics[width=0.28\textwidth]{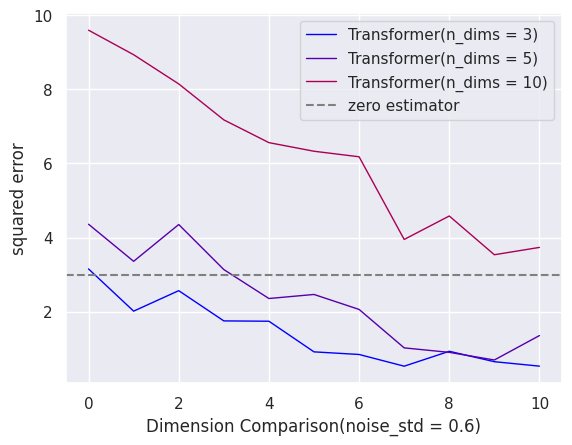}
         &
        \includegraphics[width=0.28\textwidth]{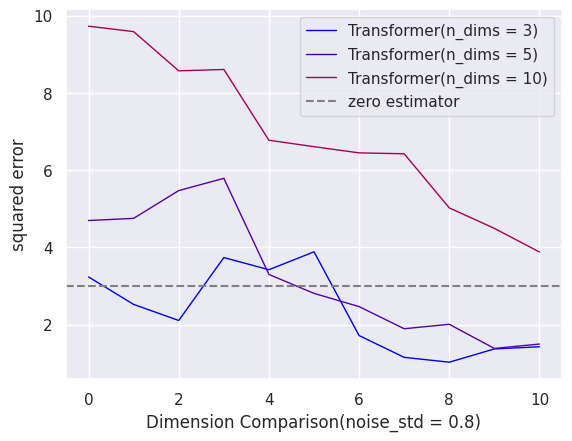} 
         & \includegraphics[width=0.28\textwidth]{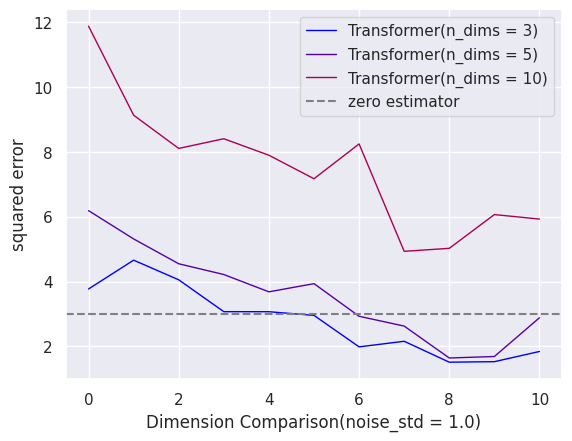}
         \\
         (d) $\sigma_{\text{test}}=0.6$ &
        (e) $\sigma_{\text{test}}=0.8$ &
        (f) $\sigma_{\text{test}}=1.0$        \\
    \end{tabular}
    \caption{\textbf{ICL performance} comparison for models with \textbf{different input dimensions}. {Each figure represents an inference noise level $\sigma_{test}$, and each line represents a model. The X-axis represents the number of in-context examples.}}
    \label{fig dim 1}
\end{figure}

\noindent\textbf{Robustness Analysis.} 
The experiment results are shown in Figure~\ref{fig dim 1}, where we can see the larger the input dimension, the worse the performance of the model in in-context learning. The performance difference between the cases of $d = 3$ and $d = 5$ is relatively small, while the model performance decreases significantly when the model $d = 10$. 

This may be because when the dimension increases, the domain of the function class to be learned expands greatly, which exceeds the model capacity, making the training much harder. Meanwhile, in the ICL process, the function to be learned itself becomes more complex due to the rise in the dimension of the input examples, so the model with higher dimensions learns poorly with the same number of examples.

\section{Revisiting input dimensions with noisy label training}
We also revisit the impact of the input dimensions on the robustness in the case of noisy label training. The overall experimental setup of this experiment is similar to that of section \ref{sub set 4 dim}, but it is worth noting that here we add Gaussian noise $\sigma \in \{0, 0.2, 0.4, 0.6, 0.8, 1\}$ during the training process in different input dimensions, and test the model with the same training noise in different dimensions in different test noise levels. The results are shown in Figure~\ref{fig dim 2}, where we can still observe the degradation of model performance with increasing dimensions in different training noises. Intuitively, we believe that the change rule of model performance with dimension in the case of training noise is not fundamentally different from that in the case of no training noise, and in both cases, the increase of dimension will make pre-training and ICL more difficult, which will result in the degradation of performance.

\begin{figure}[h]
    \centering
    \begin{tabular}{ccc}
        \includegraphics[width=0.28\textwidth]{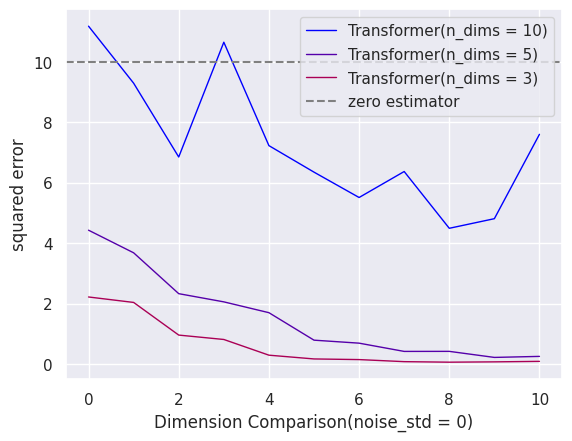}
         & \includegraphics[width=0.28\textwidth]{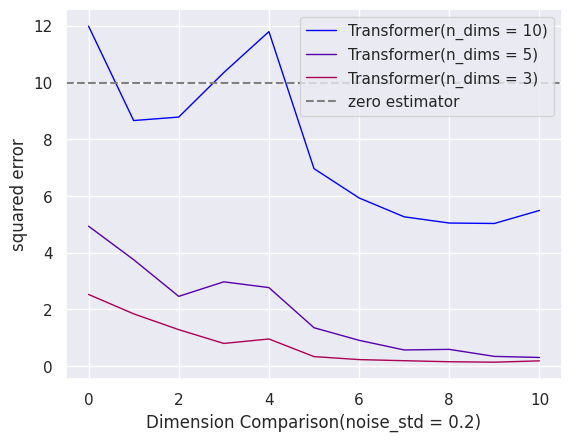}
         &
        \includegraphics[width=0.28\textwidth]{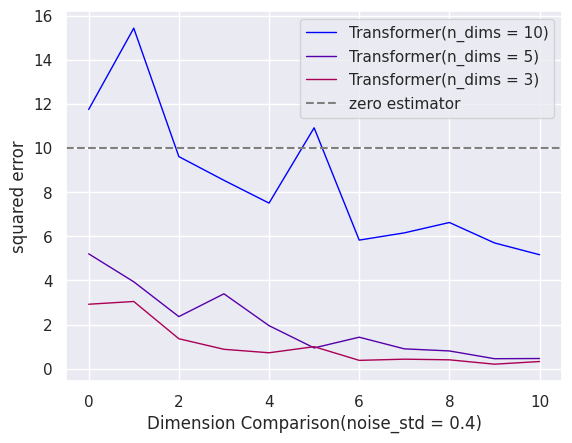}
        \\
        (a) $\sigma_{\text{test}}=0.0$ & (b) $\sigma_{\text{test}}=0.2$ &
        (c) $\sigma_{\text{test}}=0.4$ 
        \\
         \includegraphics[width=0.28\textwidth]{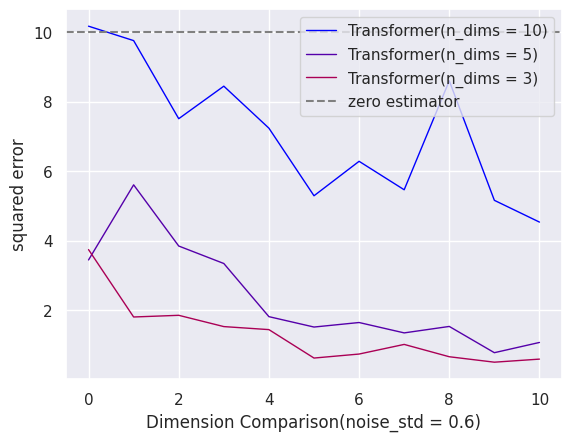}
         &
        \includegraphics[width=0.28\textwidth]{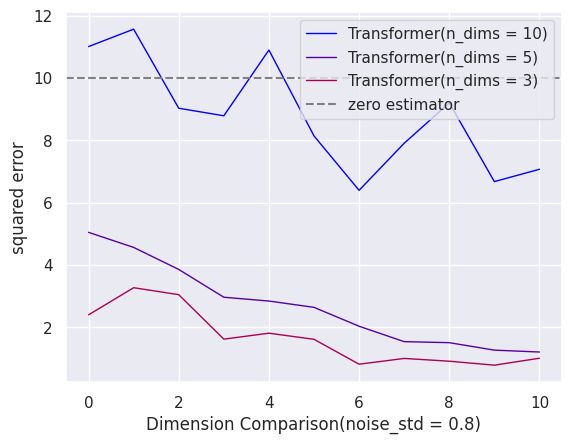} 
         & \includegraphics[width=0.28\textwidth]{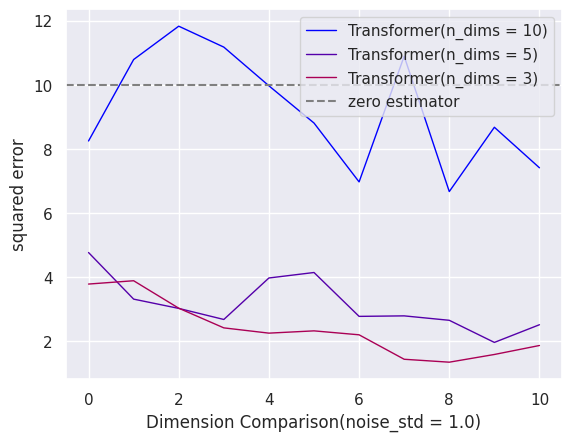}
         \\
         (d) $\sigma_{\text{test}}=0.6$ &
        (e) $\sigma_{\text{test}}=0.8$ &
        (f) $\sigma_{\text{test}}=1.0$        \\
    \end{tabular}
    \caption{\textbf{(a-h) noisy ICL performance} comparison for models trained with different dimensions. {Each figure represents an inference noise level $\sigma_{test}$, and each line represents a model trained with the same noise $\sigma = 0.6$ and different input dimension $d \in \{3,5,10\}$. The X-axis represents the number of in-context examples.}}
    \label{fig dim 2}
\end{figure}

\end{document}